\documentclass[letterpaper]{article}

\usepackage[preprint]{aaai2027}

\usepackage[hyphens]{url}
\usepackage{graphicx}
\usepackage{natbib}
\usepackage{caption}
\usepackage{algorithm}
\usepackage{algorithmic}

\usepackage[utf8]{inputenc}
\usepackage[T1]{fontenc}
\usepackage{booktabs}
\usepackage{nicefrac}
\usepackage{microtype}
\usepackage{xcolor}

\usepackage{threeparttable}
\usepackage{multirow}
\usepackage{newtxmath}
\usepackage{rotating}
\usepackage{ragged2e}
\usepackage{subcaption}

\usepackage{fontawesome5}
\usepackage{academicons}

\usepackage{listings}
\usepackage{makecell}
\usepackage[table]{xcolor}

\usepackage[capitalize,noabbrev]{cleveref}

\usepackage{array}
\usepackage{colortbl}
\usepackage{tabularx}
\usepackage{pifont}
\usepackage{adjustbox}
\usepackage{arydshln}
\usepackage{svg}
\usepackage{siunitx}

\newcommand{\cmark}{\ding{51}}
\newcommand{\xmark}{\ding{55}}

\definecolor{pltblue}{RGB}{174,199,232}
\definecolor{pltorange}{RGB}{255,152,67}
\definecolor{pltgreen}{RGB}{204,229,204}
\definecolor{pltred}{RGB}{229,204,204}
\definecolor{pltpurple}{RGB}{239,218,230}

\definecolor{tabblue}{HTML}{1f77b4}
\definecolor{taborange}{HTML}{ff7f0e}
\definecolor{tabgreen}{HTML}{2ca02c}
\definecolor{tabred}{HTML}{d62728}
\definecolor{tabpurple}{HTML}{9467bd}
\definecolor{tabpink}{HTML}{ff0080}

\definecolor{cblue}{RGB}{173,201,233}
\definecolor{clblue}{RGB}{222,234,246}
\definecolor{corange}{RGB}{255,152,67}
\definecolor{lorgange}{RGB}{255,221,149}
\definecolor{tablegray}{RGB}{180,180,180}

\usepackage{newfloat}

\DeclareCaptionStyle{ruled}{
    labelfont=normalfont,
    labelsep=colon,
    strut=off
}

\floatstyle{ruled}
\newfloat{listing}{tb}{lst}{}
\floatname{listing}{Listing}

\title{
Perturbation-Regularized Open-Vocabulary Remote Sensing Segmentation
with Unified Multi-Domain Evaluation
}

\author{
Bingyu Li\textsuperscript{\rm 1,\rm 3},
Tao Huo\textsuperscript{\rm 2,\rm 3},
Haocheng Dong\textsuperscript{\rm 1,\rm 3},
Da Zhang\textsuperscript{\rm 2,\rm 3},\\
Zhiyuan Zhao\textsuperscript{\rm 3},
Junyu Gao\textsuperscript{\rm 2,\rm 3},
Xuelong Li\textsuperscript{\rm 3}\corresponding
}

\affiliations{
\textsuperscript{\rm 1}
Department of Electronic Engineering and Information Science,
University of Science and Technology of China, China\\

\textsuperscript{\rm 2}
School of Artificial Intelligence, OPtics and ElectroNics (iOPEN),
Northwestern Polytechnical University, China\\

\textsuperscript{\rm 3}
Institute of Artificial Intelligence (TeleAI),
China Telecom, P. R. China\\[2pt]

libingyu0205@mail.ustc.edu.cn,
haocheng-dong@mail.ustc.edu.cn,\\
huotao@mail.nwpu.edu.cn,
dazhang@mail.nwpu.edu.cn,
gjy3035@gmail.com,\\
tuzixini@163.com,
xuelong\_li@ieee.org
}

\begin{document}

\maketitle

\begin{abstract}
Open-vocabulary remote sensing image segmentation (OVRSIS) aims to segment text-specified categories beyond a fixed label space. Its key challenge is to maintain reliable pixel--text correspondence under substantial appearance variation in remote-sensing imagery. Existing methods often rely on specialized vision--language representations or complex adaptation mechanisms, yet still perform deterministic matching between fixed text prototypes and individual visual features. We instead formulate OVRSIS as a \emph{distributional pixel--text alignment} problem and propose \textbf{Pi-Seg}, a lightweight perturbation-based framework. Pi-Seg introduces learnable variations into textual prototypes and dense visual features before alignment. Segmentation supervision encourages perturbations that enlarge the target-to-distractor margin while suppressing harmful feature shifts. Pi-Seg consistently improves performance under the existing \textit{single-source} OVRSISBench protocols.
To evaluate OVRSIS beyond source-specific training settings, we further introduce a unified \textit{multi-source} protocol and construct \textbf{GlobalRSOV95K}, containing approximately 95K densely annotated images and 35 semantic categories. Models are trained on this shared dataset and evaluated on an image-disjoint suite of 10 downstream datasets. The complete benchmark covers more than 170K images and 122 categories. Experiments under both single-source and multi-source protocols show that Pi-Seg improves cross-dataset transfer and generalizes effectively to practical geospatial segmentation tasks, while GlobalRSOV95K provides a common foundation for transferable OVRSIS research. \footnote{\url{https://github.com/LiBingyu01/Pi-Seg}}
\end{abstract}

\section{Introduction}
\label{sec:intro}

Open-vocabulary segmentation (OVS) extends semantic segmentation beyond a predefined label set by matching dense visual representations with textual concepts~\cite{radford2021learning,xu2022simple,cho2024cat}. This capability is particularly important in remote sensing (RS), where practical applications frequently involve categories that are absent during training or difficult to annotate exhaustively. Open-vocabulary remote sensing image segmentation (OVRSIS) therefore aims to localize text-specified geospatial categories without restricting inference to a fixed training vocabulary~\cite{li2026exploring,cao2025open,ye2025towards}.

\begin{figure}[tb]
    \centering
    \includegraphics[width=\linewidth]{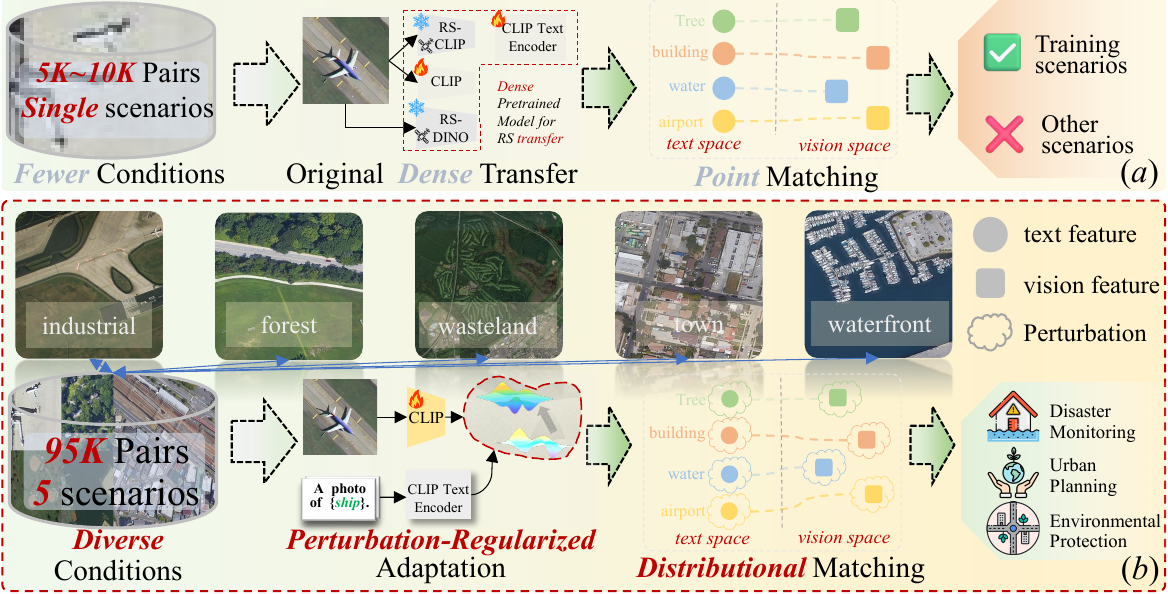}
    \caption{\textbf{Overview of the proposed study.}
    (a) Single-source deterministic (point) matching.
    (b) Multi-source distributional matching with Pi-Seg and multi-source dataset.}
    \label{fig:fig_PAPER_1TO2}
    \vspace{-1.3em}
\end{figure}

A central challenge of OVRSIS is to maintain reliable pixel--text correspondence under substantial appearance variation. Models trained on a single source distribution may therefore capture source-specific textures, scales, or contextual statistics rather than transferable semantics.
As illustrated in \cref{fig:fig_PAPER_1TO2}(a), existing methods address this issue through specialized remote-sensing vision--language representations, additional pretrained encoders, or elaborate adaptation mechanisms~\cite{li2026exploring,cao2025open,ye2025towards,wang2025rsclip}. However, segmentation is still commonly based on deterministic matching between fixed textual prototypes and individual visual features. Such point-wise matching poorly captures intra-class appearance variation and may overfit source-specific visual statistics, limiting transfer to unseen categories and domains\cite{chen2025sam}.

The upper part of \cref{fig:fig_PAPER_1TO2} also highlights a complementary limitation of existing evaluation. We therefore formulate OVRSIS as a \emph{distributional pixel--text alignment} problem and propose \textbf{Pi-Seg} (\textbf{P}erturbation-\textbf{i}njected \textbf{Seg}mentation), a lightweight framework that regularizes dense vision--language correspondence through learnable feature variations. As shown in \cref{fig:fig_PAPER_1TO2}(b), Pi-Seg provides lightweight remote-sensing adaptation by introducing learnable perturbations into textual prototypes and dense visual features before cost construction. The perturbations are optimized jointly with segmentation supervision: unfavorable shifts are suppressed, whereas variations that enlarge the target-to-distractor margin are encouraged. Pi-Seg thus extends deterministic point matching to a learned semantic neighborhood without requiring additional encoders or introducing inference-time computation.

We first evaluate Pi-Seg under the established OVRSISBench protocols, which use DLRSD or iSAID as individual training sources and transfer the learned models to downstream remote-sensing datasets. Pi-Seg achieves consistent improvements across source datasets and backbone scales. These results provide direct evidence for the effectiveness and generality of the proposed alignment strategy.
We further consider a broader multi-source setting, not as a replacement for OVRSISBench, but as a complementary protocol for evaluating models from a shared and more diverse training foundation.
To support more diverse multi-domain evaluation, we construct \textbf{GlobalRSOV95K}, a multi-domain OVRSIS training dataset containing approximately 95K densely annotated images and 35 semantic categories. It consolidates diverse remote-sensing sources into five representative environments: \emph{town}, \emph{industrial}, \emph{forest}, \emph{waterfront}, and \emph{wasteland}. Using GlobalRSOV95K as a common training source, we establish an image-disjoint suite of 10 downstream datasets. The complete benchmark covers more than 170K images and 122 categories, including satellite, aerial, and Unmanned Aerial Vehicle(UAV) imagery, as well as practical tasks such as building extraction, road extraction, and flood detection.

Together, the single-source and multi-source protocols evaluate Pi-Seg under complementary training distributions. Its consistent improvements across both settings indicate that the learned alignment is not tied to a particular source dataset, category configuration, or evaluation protocol. Meanwhile, GlobalRSOV95K provides a shared foundation for controlled cross-dataset and cross-task evaluation.

Our contributions are summarized as follows:

(1) We propose \textbf{Pi-Seg}, a lightweight perturbation-based framework that extends deterministic pixel--text matching with learnable semantic-neighborhood regularization. Pi-Seg improves dense alignment without additional inference-time computation and performs consistently under the established single-source OVRSISBench protocols.

(2) We construct \textbf{GlobalRSOV95K}, a multi-domain training dataset with approximately 95K images and 35 categories, and introduce a unified multi-source protocol with an image-disjoint downstream suite covering more than 170K images, 122 categories, and 10 datasets.

(3) We conduct evaluations under both single-source and multi-source protocols, together with cross-task transfer, demonstrating the generality of Pi-Seg.

\section{Related Work}
\label{sec:related_works}

\subsection{Semantic and Open-Vocabulary Segmentation}

Semantic segmentation has evolved from CNN-based models such as FCN~\cite{long2015fully} and DeepLab~\cite{chen2017deeplab} to Transformer-based architectures including SegFormer~\cite{xie2021segformer} and Mask2Former~\cite{cheng2022masked}. However, conventional methods remain restricted to predefined categories. Open-vocabulary segmentation (OVS) overcomes this limitation by matching dense visual features with text embeddings from vision--language models such as CLIP~\cite{radford2021learning}. Existing approaches improve dense alignment through mask prediction~\cite{ding2022open}, query matching~\cite{yu2023convolutions}, and cost aggregation~\cite{cho2024cat,xie2024sed,li2025maris, li2025exploringu}, while training-free methods such as ProxyCLIP~\cite{lan2024proxyclip} and SCLIP~\cite{wang2024sclip} adapt frozen VLMs without additional training.

\subsection{Open-Vocabulary Remote Sensing Segmentation}

OVRSIS applies OVS to geospatial imagery, where scale variation and domain shifts weaken the transferability of natural-image vision--language representations. Existing methods address this issue through knowledge transfer~\cite{cao2025open}, specialist--generalist learning~\cite{ye2025towards}, prompt adaptation~\cite{zhang2025tpov}, context modeling~\cite{huang2025score}, or additional remote-sensing priors~\cite{li2025segearth,wang2025rsclip}. Pi-Seg instead regularizes the dense pixel--text matching space directly by learning task-favorable variations around visual features and textual prototypes, without increasing inference-time computation.

\subsection{Positive-Incentive Noise}
Li~\cite{li2022positive} introduced positive-incentive noise, showing that properly optimized perturbations can benefit model learning. Zhang et al.~\cite{zhang2025variational} further developed a variational framework for learning such perturbations, while Huang et al.~\cite{huang2025enhance} applied them to global image--text alignment for few-shot classification. In contrast, Pi-Seg targets dense OVS by perturbing textual prototypes and pixel-level visual features before dense alignment. 




\section{Pi-Seg: Distributional Pixel--Text Alignment}
\label{sec:method_pi_seg}

\begin{figure*}[t]
    \centering
    \includegraphics[width=0.95\linewidth]{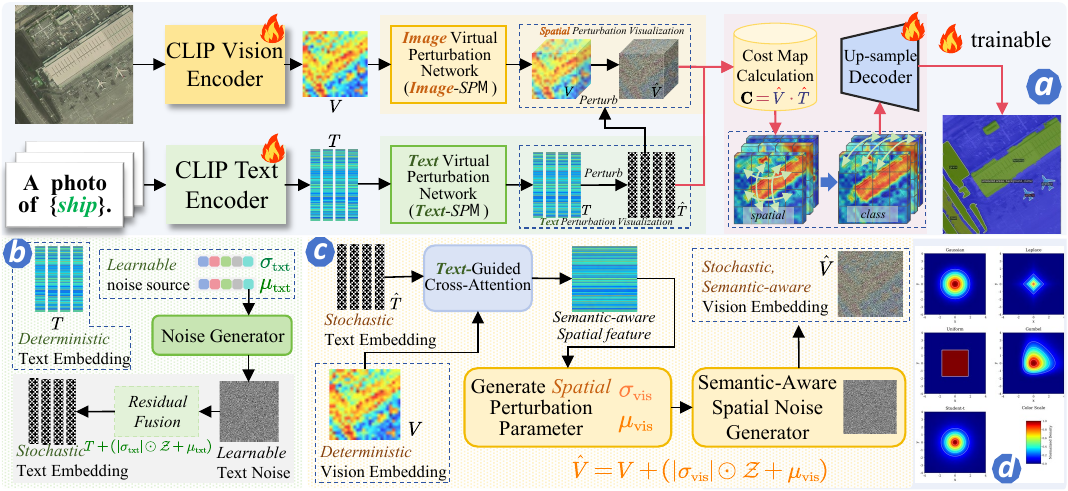}
    \caption{
    \textbf{Overview of Pi-Seg.}
    \textbf{(a)} Main pipeline.
    \textbf{(b)} Text-SPM.
    \textbf{(c)} Image-SPM.
    \textbf{(d)} Different perturbation distributions.
    }
    \label{fig:pic_pi_seg}
    \vspace{-1em}
\end{figure*}

We propose \textbf{Pi-Seg} (\textbf{P}erturbation-\textbf{i}njected \textbf{Seg}mentation), a lightweight framework that extends deterministic pixel--text matching to a learned neighborhood of semantic variations. As shown in \cref{fig:pic_pi_seg}, Pi-Seg perturbs both textual prototypes and dense visual features before cost construction. Text-SPM expands the neighborhood of category prototypes, while Image-SPM predicts spatially varying visual perturbations conditioned on text semantics. The two modules are optimized jointly by segmentation supervision, enabling robust alignment without additional encoders or inference-time computation.

\begin{figure*}[t]
    \centering
    \includegraphics[width=0.95\linewidth]{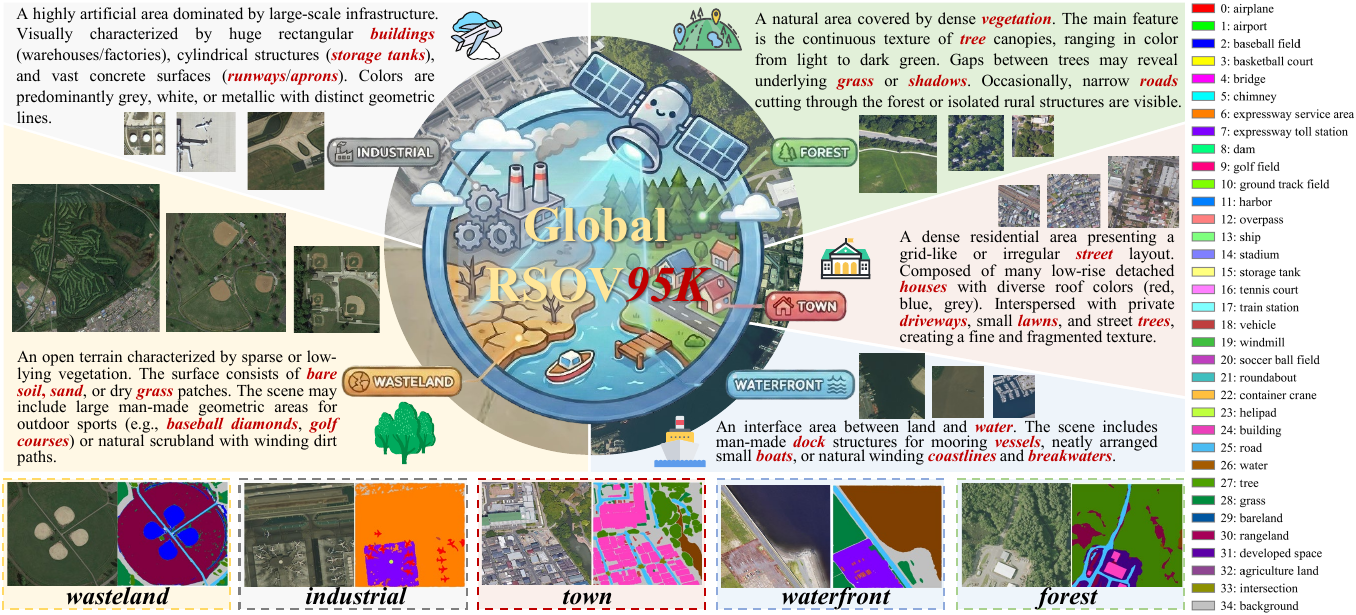}
    \caption{\textbf{Overview of GlobalRSOV95K.} The dataset provides a shared multi-source training foundation.}
    \label{fig:pic_ovrsis95k}
    \vspace{-1em}
\end{figure*}

\subsection{Overall Framework}

Given an image and a category set, frozen CLIP encoders extract dense visual features
$V\in\mathbb{R}^{HW\times C}$ and text embeddings
$T\in\mathbb{R}^{N\times C}$, where $HW$ is the number of spatial locations, $N$ is the number of categories, and $C$ is the feature dimension. Pi-Seg perturbs both modalities as
$\hat{T}=T+\epsilon_{txt}, \hat{V}=V+\epsilon_{vis}$,
where $\epsilon_{txt}$ and $\epsilon_{vis}$ are generated by Text-SPM and Image-SPM, respectively. The perturbed features are then used for cost construction, aggregation, and segmentation decoding.

\subsection{Semantic Perturbation Modules}

\textbf{Text-SPM.}
As shown in \cref{fig:pic_pi_seg}(b), Text-SPM converts deterministic category embeddings into stochastic semantic prototypes:
\begin{equation}
    \hat{T}=T+\epsilon_{txt}, \qquad
    \epsilon_{txt}=|\sigma_{txt}|\odot\mathcal{Z}+\mu_{txt},
    \label{eq:text_SPM}
\end{equation}
where $\mu_{txt},\sigma_{txt}\in\mathbb{R}^{C}$ are learnable perturbation parameters and $\mathcal{Z}$ is sampled from a predefined distribution. The mean term learns a semantic displacement, while the scale term controls the perturbation range around the original text prototype.

\textbf{Image-SPM.}
As shown in \cref{fig:pic_pi_seg}(c), Image-SPM generates semantic-aware spatial perturbations for dense visual features. It uses the perturbed text embeddings $\hat{T}$ as semantic guidance and predicts location-dependent perturbation parameters:
\begin{equation}
    \hat{V}=V+\epsilon_{vis}, \qquad
    \epsilon_{vis}=\sigma_{vis}\odot\mathcal{Z}+\mu_{vis},
    \label{eq:image_SPM}
\end{equation}
where $\mu_{vis},\sigma_{vis}\in\mathbb{R}^{HW\times C}$. Different perturbation distributions can be adopted for $\mathcal{Z}$, as illustrated in \cref{fig:pic_pi_seg}(d).

\subsection{Dense Cost Aggregation and Training Objective}

Using the perturbed features, Pi-Seg constructs the dense pixel--text cost map:
\begin{equation}
    \hat{C}(i,j)=
    \frac{\hat{V}_i^\top \hat{T}_j}
    {\|\hat{V}_i\|_2\|\hat{T}_j\|_2},
    \label{eq:cost_map}
\end{equation}
where $i$ indexes a spatial location and $j$ indexes a semantic category. The cost map is refined and decoded as
\begin{equation}
    \hat{Z}=\mathcal{D}_{\theta}\big(\mathcal{F}_{agg}(\hat{C})\big),
    \label{eq:aggregation_decoder}
\end{equation}
where $\mathcal{F}_{agg}$ denotes cost aggregation and $\mathcal{D}_{\theta}$ denotes the decoder. Cost-aggregation preliminaries are summarized in Appendix~\ref{sec:suppl_cost_aggregation}.

The model is trained with the pixel-wise segmentation loss:
\begin{equation}
    \mathcal{L}_{seg}
    =
    -\frac{1}{|\Omega|}
    \sum_{i\in\Omega}
    \log
    \frac{\exp(\hat{Z}_{i,y_i})}
    {\sum_{j=1}^{N}\exp(\hat{Z}_{i,j})},
    \label{eq:seg_loss}
\end{equation}
where $\Omega$ is the set of valid pixels and $y_i$ is the ground-truth label of pixel $i$.

\subsection{Positive-Incentive View of Pi-Seg}
\label{sec:positive_incentive_view}

Pi-Seg can be interpreted as learning a task-favorable perturbation distribution in the dense vision--language feature space. Let $\epsilon=(\epsilon_{vis},\epsilon_{txt})$ denote the joint perturbation generated by Image-SPM and Text-SPM. The perturbation modules define
\begin{equation}
    \epsilon \sim q_{\phi,\psi}(\epsilon \mid V,T),
    \qquad
    \hat{V}=V+\epsilon_{vis},
    \qquad
    \hat{T}=T+\epsilon_{txt},
    \label{eq:piseg_perturb_distribution}
\end{equation}
where $\phi$ and $\psi$ denote the parameters of Image-SPM and Text-SPM, respectively. Given the perturbed cost map $\hat{C}$ constructed from $\hat{V}$ and $\hat{T}$, Pi-Seg optimizes
\begin{equation}
    \min_{\theta,\phi,\psi}
    \mathbb{E}_{(I,Y)}
    \mathbb{E}_{\epsilon\sim q_{\phi,\psi}}
    \left[
    \mathcal{L}_{seg}
    \left(
    \mathcal{D}_{\theta}
    \big(
    \mathcal{F}_{agg}(\hat{C})
    \big),
    Y
    \right)
    \right].
    \label{eq:piseg_expected_objective}
\end{equation}
Therefore, the perturbation distribution is optimized by the segmentation objective instead of being used as random noise.

To connect this perturbation with the loss, we rewrite the pixel-wise cross-entropy from a margin view. For pixel $i$, define the target-vs-distractor margin as
\begin{equation}
    M_i(Z)=
    Z_{i,y_i}
    -
    \log
    \sum_{n\neq y_i}
    \exp(Z_{i,n}),
    \label{eq:piseg_margin}
\end{equation}
where $Z_{i,y_i}$ is the logit of the ground-truth category and the second term aggregates the competing categories. The pixel-wise cross-entropy can then be written as
\begin{equation}
    \ell_i(Z,Y)
    =
    \log
    \left(
    1+\exp(-M_i(Z))
    \right).
    \label{eq:piseg_margin_loss}
\end{equation}
This formulation shows that increasing the target-vs-distractor margin directly reduces the pixel-wise segmentation loss.

Accordingly, a perturbation is regarded as positive-incentive when it increases the expected margin after perturbation:
\begin{equation}
    \mathbb{E}_{q_{\phi,\psi}}
    \left[
    M_i(\hat{Z})-M_i(Z)
    \right]
    >0 .
    \label{eq:piseg_positive_margin}
\end{equation}
This condition links the learned perturbation to the optimization objective. Perturbations that damage dense pixel--text alignment increase the segmentation loss and are suppressed during training, while perturbations that enlarge the target-vs-distractor margin are encouraged. The full derivation is provided in Appendix~\ref{sec:suppl_variational_piseg}.

\section{Unified Multi-Source Evaluation}
\label{sec:globalrsov95k_benchmark}

We further consider a unified multi-source setting. The goal is to train all methods on the same diverse observation base and evaluate them under identical downstream conditions. 
We therefore construct \textbf{GlobalRSOV95K} as a common multi-domain training dataset and build an image-disjoint down-stream suite for controlled transfer evaluation.

\begin{figure}[tb]
    \centering
    \includegraphics[width=\linewidth]{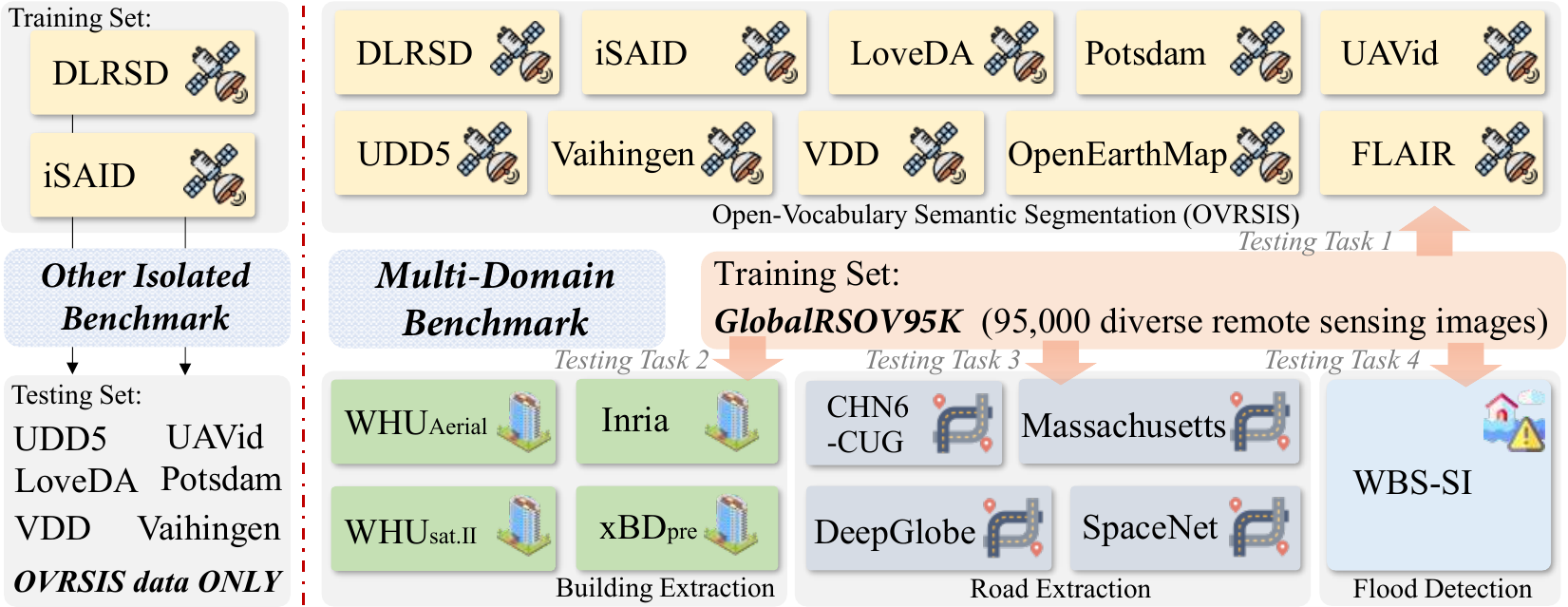}
    \caption{\textbf{Unified multi-source OVRSIS protocol.}}
    \label{fig:pic_fig_2_ovrsisv2}
    \vspace{-0.8em}
\end{figure}

\begin{figure}[tb]
    \centering
    \includegraphics[width=\linewidth]{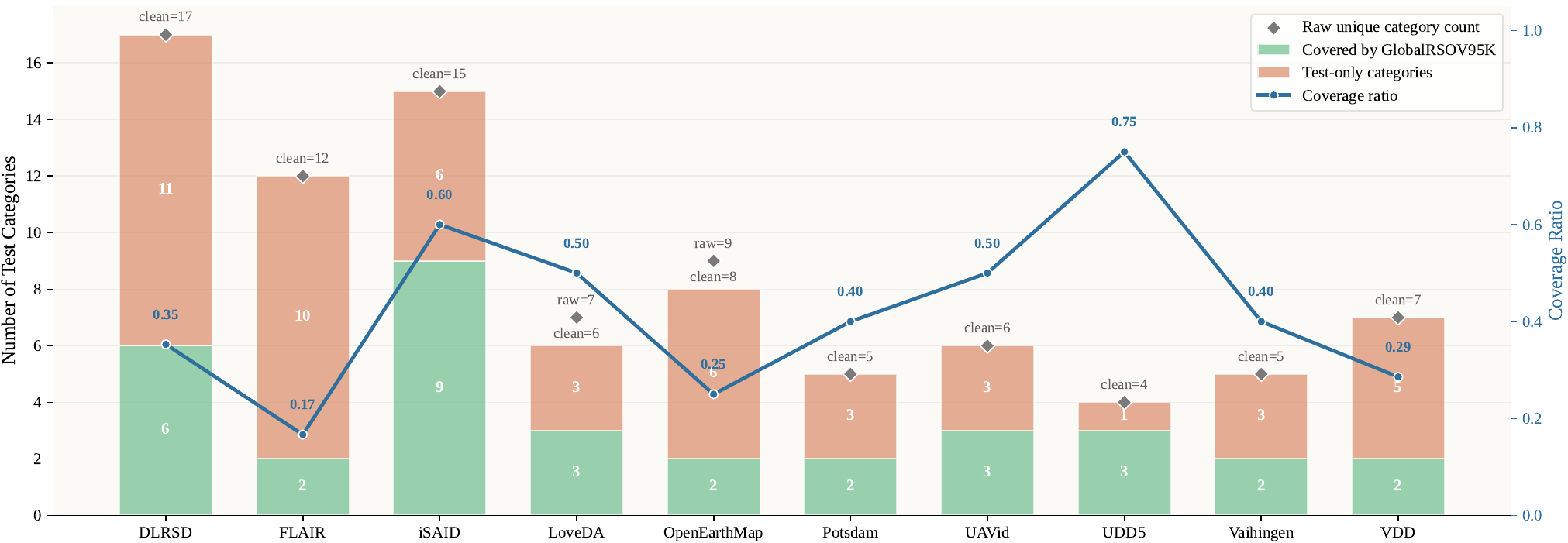}
    \caption{\textbf{Category overlap under the unified protocol.}}
    \label{fig:fig_test_centric_overlap_bar_GlobalRSOV95K}
    \vspace{-0.6em}
\end{figure}

\begin{table*}[htbp]
    \centering
    \renewcommand{\arraystretch}{1.0}
    \setlength{\tabcolsep}{1.3pt}
    \resizebox{\linewidth}{!}{
    \begin{tabular}{lcccccccccccccccccccc}
        \specialrule{0.95pt}{2pt}{2pt}

        \multicolumn{21}{c}{\textbf{DLRSD as Training Dataset}} \\
        \specialrule{0.95pt}{2pt}{2pt}

        \multirow{2}*{\textbf{Method}}
        & \multirow{2}*{\textbf{Backbone}}
        & \multirow{2}*{\textbf{Type}}
        & \multicolumn{2}{c}{\textbf{DLRSD}}
        & \multicolumn{2}{c}{\textbf{iSAID}}
        & \multicolumn{2}{c}{\textbf{LoveDA}}
        & \multicolumn{2}{c}{\textbf{Potsdam}}
        & \multicolumn{2}{c}{\textbf{UAVid}}
        & \multicolumn{2}{c}{\textbf{UDD5}}
        & \multicolumn{2}{c}{\textbf{Vaihingen}}
        & \multicolumn{2}{c}{\textbf{VDD}}
        & \multicolumn{2}{c}{\textbf{Mean w/o DLRSD}} \\

        \cmidrule(lr){4-5}
        \cmidrule(lr){6-7}
        \cmidrule(lr){8-9}
        \cmidrule(lr){10-11}
        \cmidrule(lr){12-13}
        \cmidrule(lr){14-15}
        \cmidrule(lr){16-17}
        \cmidrule(lr){18-19}
        \cmidrule(lr){20-21}

        & & &
        mIoU & mACC &
        mIoU & mACC &
        mIoU & mACC &
        mIoU & mACC &
        mIoU & mACC &
        mIoU & mACC &
        mIoU & mACC &
        mIoU & mACC &
        m-mIoU & m-mACC \\

        \midrule


        OVRS\cite{cao2025open}%
        $_{\textcolor{gray}{\text{\textit{TGRS2025}}}}$
        & ViT-B & OVRSIS
        & -- & --
        & 39.09 & 54.43
        & 28.67 & 52.10
        & 27.47 & 42.07
        & 25.23 & 40.18
        & 39.10 & 50.65
        & 33.71 & 49.01
        & 37.34 & 55.15
        & 32.94 & 49.08 \\

        GSNet\cite{ye2025towards}%
        $_{\textcolor{gray}{\text{\textit{AAAI2025}}}}$
        & ViT-B & OVRSIS
        & -- & --
        & \underline{42.00} & \underline{59.19}
        & 29.32 & 53.02
        & 26.46 & 43.20
        & 25.42 & 40.70
        & 40.05 & 51.72
        & \underline{35.15} & \underline{52.62}
        & 38.10 & 56.01
        & 33.79 & 50.92 \\

        RSKT-Seg\cite{li2026exploring}%
        $_{\textcolor{gray}{\text{\textit{AAAI2026}}}}$
        & ViT-B & OVRSIS
        & -- & --
        & \textbf{44.04} & \textbf{61.23}
        & \underline{32.49} & \underline{55.67}
        & \textbf{34.53} & \textbf{50.71}
        & \underline{28.14} & \underline{42.86}
        & \underline{42.99} & \underline{57.81}
        & \textbf{37.16} & \textbf{54.81}
        & \underline{41.22} & \underline{58.09}
        & \underline{37.22} & \underline{54.45} \\

        Pi-Seg
        & ViT-B & OVRSIS
        & -- & --
        & 40.70 & 56.53
        & \textbf{37.43} & \textbf{64.47}
        & \underline{28.07} & \underline{46.88}
        & \textbf{34.45} & \textbf{50.94}
        & \textbf{52.47} & \textbf{67.30}
        & 30.53 & 49.82
        & \textbf{44.15} & \textbf{63.72}
        & \textbf{38.26} & \textbf{57.09} \\


        \midrule

        OVRS\cite{cao2025open}%
        $_{\textcolor{gray}{\text{\textit{TGRS2025}}}}$
        & ViT-L & OVRSIS
        & -- & --
        & 52.65 & 69.59
        & 31.53 & 59.82
        & 36.44 & 50.17
        & 24.13 & 34.83
        & 40.82 & 54.24
        & 43.50 & \underline{63.31}
        & 37.23 & 56.34
        & 38.04 & 55.47 \\

        GSNet\cite{ye2025towards}%
        $_{\textcolor{gray}{\text{\textit{AAAI2025}}}}$
        & ViT-L & OVRSIS
        & -- & --
        & \underline{53.73} & \underline{71.57}
        & 32.52 & \underline{60.23}
        & \underline{37.85} & \underline{52.35}
        & 24.22 & 35.03
        & 40.92 & \underline{57.04}
        & \underline{44.13} & 62.38
        & 37.34 & 57.04
        & 38.67 & 56.52 \\

        RSKT-Seg\cite{li2026exploring}%
        $_{\textcolor{gray}{\text{\textit{AAAI2026}}}}$
        & ViT-L & OVRSIS
        & -- & --
        & \textbf{54.32} & \textbf{71.72}
        & \underline{33.23} & 57.41
        & \textbf{38.44} & \textbf{56.69}
        & \underline{25.72} & \underline{40.89}
        & \underline{42.10} & 56.39
        & 42.69 & 63.29
        & \underline{39.69} & \underline{58.60}
        & \underline{39.46} & \underline{57.86} \\

        Pi-Seg
        & ViT-L & OVRSIS
        & -- & --
        & 53.50 & 69.68
        & \textbf{39.00} & \textbf{61.76}
        & 35.33 & 48.33
        & \textbf{33.58} & \textbf{48.78}
        & \textbf{59.08} & \textbf{72.58}
        & \textbf{44.18} & \textbf{64.74}
        & \textbf{47.73} & \textbf{67.10}
        & \textbf{44.63} & \textbf{61.85} \\


        \specialrule{0.95pt}{2pt}{2pt}
        \multicolumn{21}{c}{\textbf{iSAID as Training Dataset}} \\
        \specialrule{0.95pt}{2pt}{2pt}

        \multirow{2}*{\textbf{Method}}
        & \multirow{2}*{\textbf{Backbone}}
        & \multirow{2}*{\textbf{Type}}
        & \multicolumn{2}{c}{\textbf{DLRSD}}
        & \multicolumn{2}{c}{\textbf{iSAID}}
        & \multicolumn{2}{c}{\textbf{LoveDA}}
        & \multicolumn{2}{c}{\textbf{Potsdam}}
        & \multicolumn{2}{c}{\textbf{UAVid}}
        & \multicolumn{2}{c}{\textbf{UDD5}}
        & \multicolumn{2}{c}{\textbf{Vaihingen}}
        & \multicolumn{2}{c}{\textbf{VDD}}
        & \multicolumn{2}{c}{\textbf{Mean w/o iSAID}} \\

        \cmidrule(lr){4-5}
        \cmidrule(lr){6-7}
        \cmidrule(lr){8-9}
        \cmidrule(lr){10-11}
        \cmidrule(lr){12-13}
        \cmidrule(lr){14-15}
        \cmidrule(lr){16-17}
        \cmidrule(lr){18-19}
        \cmidrule(lr){20-21}

        & & &
        mIoU & mACC &
        mIoU & mACC &
        mIoU & mACC &
        mIoU & mACC &
        mIoU & mACC &
        mIoU & mACC &
        mIoU & mACC &
        mIoU & mACC &
        m-mIoU & m-mACC \\

        \midrule


        OVRS\cite{cao2025open}%
        $_{\textcolor{gray}{\text{\textit{TGRS2025}}}}$
        & ViT-B & OVRSIS
        & 21.06 & 45.48
        & -- & --
        & 25.30 & 42.32
        & 15.57 & \underline{38.94}
        & \underline{16.22} & 24.05
        & 11.90 & 38.80
        & \underline{14.66} & 38.68
        & 21.42 & \textbf{51.25}
        & 18.02 & 39.93 \\

        GSNet\cite{ye2025towards}%
        $_{\textcolor{gray}{\text{\textit{AAAI2025}}}}$
        & ViT-B & OVRSIS
        & \textbf{26.20} & \textbf{57.07}
        & -- & --
        & 26.80 & 42.84
        & 15.12 & 36.16
        & 15.93 & 24.33
        & 12.44 & 39.62
        & 14.25 & \underline{41.15}
        & \underline{22.22} & 42.33
        & 18.99 & 40.50 \\

        RSKT-Seg\cite{li2026exploring}%
        $_{\textcolor{gray}{\text{\textit{AAAI2026}}}}$
        & ViT-B & OVRSIS
        & \underline{24.80} & \underline{55.94}
        & -- & --
        & \underline{28.07} & \underline{46.58}
        & \textbf{20.28} & \textbf{46.71}
        & \textbf{17.15} & \textbf{34.78}
        & \underline{13.94} & \underline{43.47}
        & \textbf{17.47} & \textbf{50.84}
        & \textbf{25.34} & \underline{47.18}
        & \textbf{21.01} & \textbf{46.50} \\

        Pi-Seg
        & ViT-B & OVRSIS
        & 23.62 & 50.82
        & -- & --
        & \textbf{30.90} & \textbf{50.39}
        & \underline{17.37} & 37.25
        & 12.37 & \underline{33.69}
        & \textbf{17.51} & \textbf{45.45}
        & 12.58 & 38.49
        & 20.18 & 39.67
        & \underline{19.22} & \underline{42.25} \\


        \midrule

        OVRS\cite{cao2025open}%
        $_{\textcolor{gray}{\text{\textit{TGRS2025}}}}$
        & ViT-L & OVRSIS
        & \underline{32.25} & 60.35
        & -- & --
        & 27.98 & 49.01
        & 26.39 & 50.15
        & \underline{16.24} & 27.00
        & 31.88 & 55.32
        & \textbf{28.80} & \textbf{54.20}
        & 31.01 & \underline{55.30}
        & 27.79 & 50.19 \\

        GSNet\cite{ye2025towards}%
        $_{\textcolor{gray}{\text{\textit{AAAI2025}}}}$
        & ViT-L & OVRSIS
        & 31.50 & \textbf{63.05}
        & -- & --
        & 27.21 & \underline{49.53}
        & \underline{28.50} & \textbf{52.00}
        & 15.98 & \underline{27.10}
        & 32.24 & \underline{56.20}
        & 25.10 & \underline{52.80}
        & 32.07 & 55.21
        & 27.51 & \underline{50.84} \\

        RSKT-Seg\cite{li2026exploring}%
        $_{\textcolor{gray}{\text{\textit{AAAI2026}}}}$
        & ViT-L & OVRSIS
        & 31.57 & \underline{61.66}
        & -- & --
        & \underline{29.62} & \textbf{51.01}
        & \textbf{28.57} & \underline{51.69}
        & \textbf{17.36} & \textbf{35.75}
        & \underline{34.01} & \textbf{57.38}
        & \underline{25.55} & 52.66
        & \underline{33.40} & \textbf{56.50}
        & \underline{28.58} & \textbf{52.38} \\

        Pi-Seg
        & ViT-L & OVRSIS
        & \textbf{33.67} & 59.44
        & -- & --
        & \textbf{31.45} & 44.23
        & 28.14 & 46.39
        & 12.34 & 23.57
        & \textbf{37.24} & 53.87
        & 23.83 & 47.32
        & \textbf{38.24} & 54.25
        & \textbf{29.27} & 47.01 \\

        \specialrule{0.95pt}{2pt}{2pt}
    \end{tabular}
    }

    \vspace{-0.6em}
    \caption{\textbf{Single-source OVRSIS transfer results under DLRSD and iSAID training protocols.}}
    \label{tab:01_main_table_DLRSD_iSAID}
    \vspace{-0.6em}
\end{table*}
\begin{table*}[ht]
\centering
\large
\renewcommand{\arraystretch}{1.0}
\setlength{\tabcolsep}{1.2pt}
\resizebox{\linewidth}{!}{%
\begin{tabular}{lcc*{22}{c}}
\specialrule{0.95pt}{2pt}{2pt}
\multicolumn{25}{c}{
\textbf{GlobalRSOV95K as Training Foundation}
} \\
\specialrule{0.95pt}{2pt}{2pt}

\multirow{2}*{\textbf{Method}}
& \multirow{2}*{\textbf{Backbone}}
& \multirow{2}*{\textbf{Type}}
& \multicolumn{2}{c}{\textbf{DLRSD}}
& \multicolumn{2}{c}{\textbf{FLAIR}}
& \multicolumn{2}{c}{\textbf{iSAID}}
& \multicolumn{2}{c}{\textbf{LoveDA}}
& \multicolumn{2}{c}{\textbf{OpenEarthMap}}
& \multicolumn{2}{c}{\textbf{Potsdam}}
& \multicolumn{2}{c}{\textbf{UAVid}}
& \multicolumn{2}{c}{\textbf{UDD5}}
& \multicolumn{2}{c}{\textbf{Vaihingen}}
& \multicolumn{2}{c}{\textbf{VDD}}
& \multicolumn{2}{c}{\textbf{Mean}} \\

\cmidrule(lr){4-5}
\cmidrule(lr){6-7}
\cmidrule(lr){8-9}
\cmidrule(lr){10-11}
\cmidrule(lr){12-13}
\cmidrule(lr){14-15}
\cmidrule(lr){16-17}
\cmidrule(lr){18-19}
\cmidrule(lr){20-21}
\cmidrule(lr){22-23}
\cmidrule(lr){24-25}

& & &
mIoU & mACC &
mIoU & mACC &
mIoU & mACC &
mIoU & mACC &
mIoU & mACC &
mIoU & mACC &
mIoU & mACC &
mIoU & mACC &
mIoU & mACC &
mIoU & mACC &
m-mIoU & m-mACC \\

\specialrule{0.95pt}{2pt}{2pt}


ProxyCLIP\cite{lan2024proxyclip}%
$_{\textcolor{gray}{\text{\textit{ECCV2024}}}}$
& ViT-B & OVS (TF)
& 24.03 & 43.28
& 15.85 & 30.39
& 29.01 & 38.59
& 10.02 & 14.66
& 34.70 & 54.72
& 30.85 & 51.11
& 17.40 & 26.47
& 42.89 & 62.66
& 21.24 & 40.95
& 40.86 & 57.78
& 26.68 & 42.06 \\

Trident
& ViT-B & OVS (TF)
& 26.31 & 47.54
& 18.12 & 33.67
& 28.66 & 47.93
& 10.54 & 14.99
& 33.35 & 52.65
& 31.53 & 50.03
& 14.44 & 24.36
& 41.55 & 54.64
& 21.40 & 40.58
& 39.04 & 55.09
& 26.49 & 42.15 \\

SegEarth-OV\cite{li2025segearth}%
$_{\textcolor{gray}{\text{\textit{CVPR2025}}}}$
& ViT-B & OVRSIS (TF)
& 23.76 & 41.64
& 13.09 & 26.65
& 23.36 & 32.34
& 10.93 & 15.44
& 36.97 & 59.51
& 29.40 & 50.29
& 15.99 & 25.18
& 47.69 & 63.28
& 17.12 & 37.36
& 38.62 & 58.86
& 25.69 & 41.06 \\

\specialrule{0.95pt}{2pt}{2pt}


FCCLIP\cite{yu2023convolutions}%
$_{\textcolor{gray}{\text{\textit{NeurIPS2023}}}}$
& ConvNeXt-B & OVS
& 26.48 & 45.98
& 13.92 & 24.10
& 27.61 & 49.14
& 36.00 & 55.84
& 30.13 & \textbf{56.52}
& 20.37 & 34.68
& 33.66 & 54.78
& 50.90 & \underline{73.18}
& 22.91 & 39.98
& 37.38 & 52.76
& 29.94 & 48.70 \\

CAT-SEG\cite{cho2024cat}%
$_{\textcolor{gray}{\text{\textit{CVPR2024}}}}$
& ViT-B & OVS
& 35.85 & 54.91
& 20.74 & 37.99
& \textbf{49.19} & \underline{67.81}
& 43.88 & \textbf{64.75}
& \underline{33.78} & \underline{55.37}
& \underline{30.54} & 43.31
& 37.34 & 55.58
& 52.14 & 71.82
& 30.84 & 44.25
& \textbf{44.47} & \underline{61.44}
& \underline{37.88} & \underline{55.72} \\

OVRS\cite{cao2025open}%
$_{\textcolor{gray}{\text{\textit{TGRS2025}}}}$
& ViT-B & OVRSIS
& \underline{37.30} & \underline{57.52}
& \underline{22.48} & \textbf{39.45}
& 47.62 & 66.21
& \underline{44.05} & \underline{64.54}
& \textbf{33.79} & 54.74
& 30.31 & 43.50
& \textbf{43.56} & \textbf{59.11}
& \underline{53.72} & 70.95
& 26.49 & 40.00
& 35.23 & 49.01
& 37.46 & 54.50 \\

GSNET\cite{ye2025towards}%
$_{\textcolor{gray}{\text{\textit{AAAI2025}}}}$
& ViT-B & OVRSIS
& 35.09 & 55.10
& 20.21 & 37.94
& 48.62 & 67.27
& 41.29 & 64.04
& 33.71 & 53.44
& 30.25 & 42.84
& 38.63 & 54.56
& 45.55 & 63.36
& 30.56 & 43.09
& 38.63 & 54.10
& 36.25 & 53.57 \\

RSKT-SEG\cite{li2026exploring}%
$_{\textcolor{gray}{\text{\textit{AAAI2026}}}}$
& ViT-B & OVRSIS
& 34.96 & 55.65
& 20.52 & \underline{39.19}
& \underline{49.00} & \textbf{69.52}
& 42.33 & 63.06
& 29.39 & 47.41
& \textbf{33.41} & \textbf{47.64}
& 35.04 & 49.94
& 44.27 & 59.62
& \underline{31.35} & \textbf{46.42}
& 35.96 & 52.49
& 35.62 & 53.09 \\

Pi-Seg
& ViT-B & OVRSIS
& \textbf{39.69} & \textbf{58.33}
& \textbf{22.97} & 37.48
& 47.54 & 66.95
& \textbf{45.92} & 63.56
& 32.37 & 51.93
& 30.38 & \underline{45.30}
& \underline{40.22} & \underline{57.76}
& \textbf{59.61} & \textbf{77.62}
& \textbf{31.70} & \underline{46.31}
& \underline{44.39} & \textbf{63.80}
& \textbf{39.48} & \textbf{56.90} \\

\specialrule{0.95pt}{2pt}{2pt}


FCCLIP\cite{yu2023convolutions}%
$_{\textcolor{gray}{\text{\textit{NeurIPS2023}}}}$
& ConvNeXt-L & OVS
& 26.71 & 48.21
& 20.25 & 35.12
& 29.37 & 49.09
& 37.46 & 59.82
& 32.58 & 54.89
& 22.25 & 38.62
& 34.62 & 52.76
& 42.03 & 53.58
& 23.51 & 43.82
& 30.00 & 47.69
& 29.88 & 48.36 \\

CAT-SEG\cite{cho2024cat}%
$_{\textcolor{gray}{\text{\textit{CVPR2024}}}}$
& ViT-L & OVS
& 43.89 & 62.78
& \underline{26.52} & 42.67
& 58.22 & \underline{77.71}
& 43.96 & \textbf{67.81}
& 34.97 & \underline{55.15}
& 38.03 & 52.93
& 33.49 & 50.18
& \underline{57.80} & \underline{72.60}
& 40.25 & 60.89
& 47.74 & \underline{67.34}
& 42.49 & \underline{61.01} \\

OVRS\cite{cao2025open}%
$_{\textcolor{gray}{\text{\textit{TGRS2025}}}}$
& ViT-L & OVRSIS
& 43.02 & 61.12
& 25.51 & 42.00
& \textbf{59.65} & \textbf{78.69}
& \underline{43.98} & \underline{66.84}
& 33.58 & 52.32
& \textbf{40.61} & \underline{55.86}
& \textbf{42.98} & \underline{54.24}
& 49.82 & 61.92
& 38.78 & 57.60
& 49.72 & 62.49
& \underline{42.77} & 59.31 \\

GSNET\cite{ye2025towards}%
$_{\textcolor{gray}{\text{\textit{AAAI2025}}}}$
& ViT-L & OVRSIS
& 42.88 & 61.75
& 26.25 & 43.35
& \underline{58.49} & 76.54
& \textbf{44.73} & 55.77
& \underline{35.29} & 52.28
& 33.62 & 51.16
& 32.36 & 43.99
& 56.54 & 68.56
& 36.25 & 54.94
& \textbf{52.45} & 65.86
& 41.89 & 57.42 \\

RSKT-SEG\cite{li2026exploring}%
$_{\textcolor{gray}{\text{\textit{AAAI2026}}}}$
& ViT-L & OVRSIS
& \underline{44.31} & \underline{63.67}
& 26.09 & \textbf{44.49}
& 58.15 & 77.43
& 36.71 & 63.53
& 31.18 & 50.64
& 37.11 & \textbf{57.69}
& 31.14 & 46.73
& 44.36 & 59.28
& \textbf{50.81} & \textbf{72.45}
& 41.16 & 63.45
& 40.10 & 59.94 \\

Pi-Seg
& ViT-L & OVRSIS
& \textbf{45.64} & \textbf{64.31}
& \textbf{27.37} & \underline{44.25}
& 54.95 & 76.03
& 43.26 & 66.74
& \textbf{36.23} & \textbf{57.46}
& \underline{39.07} & 53.75
& \underline{41.25} & \textbf{56.89}
& \textbf{59.60} & \textbf{74.74}
& \underline{44.94} & \underline{66.11}
& \underline{51.72} & \textbf{71.27}
& \textbf{44.40} & \textbf{63.16} \\

\specialrule{0.95pt}{2pt}{2pt}
\end{tabular}
}

\vspace{-0.5em}
\caption{\textbf{Main comparison on the unified multi-domain OVRSIS benchmark.}}
\label{tab:02_main_table_OVRSIS95K}
\vspace{-1.3em}
\end{table*}
\begin{table*}[ht]
\centering
\large
\vspace{-0.8em}
\renewcommand{\arraystretch}{1.1}
\setlength{\tabcolsep}{1pt}
\resizebox{0.99\linewidth}{!}{%
\begin{tabular}{lll*{20}{c}}
\specialrule{0.95pt}{2pt}{2pt}
\multicolumn{23}{c}{\textbf{GlobalRSOV95K as Training Foundation}} \\
\specialrule{0.95pt}{2pt}{2pt}

\multirow{3}*{\textbf{Method}}
& \multirow{3}*{\textbf{Backbone}}
& \multirow{3}*{\textbf{Type}}
& \multicolumn{8}{c}{\textbf{Building Extraction}}
& \multicolumn{8}{c}{\textbf{Road Extraction}}
& \multicolumn{2}{c}{\textbf{Flood Detection}}
& \multicolumn{2}{c}{\textbf{Mean}} \\

\cmidrule(lr){4-11}
\cmidrule(lr){12-19}
\cmidrule(lr){20-21}
\cmidrule(lr){22-23}

& & &
\multicolumn{2}{c}{WHU$^{Aerial}$}
& \multicolumn{2}{c}{WHU$^{Sat.II}$}
& \multicolumn{2}{c}{Inria}
& \multicolumn{2}{c}{xBD$^{pre}$}
& \multicolumn{2}{c}{CHN6-CUG}
& \multicolumn{2}{c}{DeepGlobe}
& \multicolumn{2}{c}{Massachusetts}
& \multicolumn{2}{c}{SpaceNet}
& \multicolumn{2}{c}{WBS-SI}
& \multicolumn{2}{c}{--} \\

\cmidrule(lr){4-5}
\cmidrule(lr){6-7}
\cmidrule(lr){8-9}
\cmidrule(lr){10-11}
\cmidrule(lr){12-13}
\cmidrule(lr){14-15}
\cmidrule(lr){16-17}
\cmidrule(lr){18-19}
\cmidrule(lr){20-21}
\cmidrule(lr){22-23}

& & &
mIoU & mACC &
mIoU & mACC &
mIoU & mACC &
mIoU & mACC &
mIoU & mACC &
mIoU & mACC &
mIoU & mACC &
mIoU & mACC &
mIoU & mACC &
m-mIoU & m-mACC \\

\specialrule{0.95pt}{2pt}{2pt}


ClearCLIP\cite{lan2024clearclip}%
$_{\textcolor{gray}{\text{\tiny\textit{ECCV2024}}}}$
& ViT-B & OVS (TF)
& 36.60 & --
& 20.80 & --
& 39.00 & --
& 30.10 & --
& 25.50 & --
& 5.70 & --
& 6.40 & --
& 16.30 & --
& 44.90 & --
& 25.03 & -- \\

SegEarth-OV\cite{li2025segearth}%
$_{\textcolor{gray}{\text{\tiny\textit{CVPR2025}}}}$
& ViT-B & OVRSIS (TF)
& 49.20 & --
& 28.40 & --
& 44.60 & --
& 37.00 & --
& 35.40 & --
& 17.80 & --
& 11.50 & --
& 23.80 & --
& 60.20 & --
& 34.21 & -- \\

\specialrule{0.95pt}{2pt}{2pt}


CAT-SEG\cite{cho2024cat}%
$_{\textcolor{gray}{\text{\tiny\textit{CVPR2024}}}}$
& ViT-B & OVS
& \underline{78.11} & \underline{95.60}
& 52.50 & 88.50
& \underline{75.87} & \underline{90.29}
& \underline{74.84} & \underline{81.51}
& 48.19 & \underline{84.91}
& 51.11 & 73.05
& 45.37 & \underline{73.13}
& 57.43 & \textbf{78.32}
& 71.91 & \underline{85.38}
& 61.70 & \textbf{83.41} \\

OVRS\cite{cao2025open}%
$_{\textcolor{gray}{\text{\tiny\textit{TGRS2025}}}}$
& ViT-B & OVRSIS
& 77.49 & 94.92
& 53.46 & 86.61
& 75.73 & 88.02
& 70.61 & 74.92
& 49.10 & 84.46
& 53.78 & 67.14
& 44.37 & 69.48
& \underline{57.81} & 70.49
& \underline{73.95} & 83.42
& 61.81 & 79.94 \\

GSNET\cite{ye2025towards}%
$_{\textcolor{gray}{\text{\tiny\textit{AAAI2025}}}}$
& ViT-B & OVRSIS
& 70.17 & 93.67
& 46.15 & 86.64
& 73.78 & 90.06
& \textbf{75.98} & \textbf{84.51}
& 46.83 & 84.78
& 53.92 & \textbf{77.03}
& 42.40 & \textbf{73.94}
& 52.73 & \underline{77.95}
& 59.12 & 82.01
& 57.90 & \underline{83.40} \\

RSKT-SEG\cite{li2026exploring}%
$_{\textcolor{gray}{\text{\tiny\textit{AAAI2026}}}}$
& ViT-B & OVRSIS
& 75.26 & 94.16
& \underline{60.75} & \textbf{91.11}
& 70.68 & 88.57
& 68.98 & 77.25
& \underline{57.30} & \textbf{85.10}
& \underline{57.34} & 71.91
& \underline{49.86} & 60.88
& 53.36 & 71.95
& 70.66 & 84.93
& \underline{62.69} & 80.65 \\

Pi-Seg
& ViT-B & OVRSIS
& \textbf{85.88} & \textbf{96.14}
& \textbf{66.16} & \underline{89.01}
& \textbf{78.61} & \textbf{90.55}
& 74.51 & 79.67
& \textbf{62.12} & 81.75
& \textbf{62.44} & \underline{73.63}
& \textbf{55.00} & 64.24
& \textbf{59.40} & 74.66
& \textbf{78.49} & \textbf{86.41}
& \textbf{69.18} & 81.78 \\

\specialrule{0.95pt}{2pt}{2pt}


CAT-SEG\cite{cho2024cat}%
$_{\textcolor{gray}{\text{\tiny\textit{CVPR2024}}}}$
& ViT-L & OVS
& \underline{80.91} & \underline{96.43}
& 56.26 & 92.25
& 75.11 & \underline{91.15}
& 75.23 & \textbf{92.15}
& 54.07 & 88.76
& 56.10 & 80.29
& 49.82 & \underline{74.29}
& 57.29 & \textbf{81.73}
& \underline{76.56} & \underline{88.54}
& 64.59 & \underline{87.29} \\

OVRS\cite{cao2025open}%
$_{\textcolor{gray}{\text{\tiny\textit{TGRS2025}}}}$
& ViT-L & OVRSIS
& 79.86 & 95.88
& 54.60 & 91.53
& \underline{76.17} & 89.96
& 70.99 & 76.46
& 56.00 & 88.81
& 58.82 & 71.32
& 50.83 & 65.73
& 55.30 & 80.86
& 75.78 & 86.19
& 64.26 & 82.97 \\

GSNET\cite{ye2025towards}%
$_{\textcolor{gray}{\text{\tiny\textit{AAAI2025}}}}$
& ViT-L & OVRSIS
& 79.92 & 96.24
& \underline{58.57} & \underline{92.66}
& 72.12 & 90.69
& \underline{76.23} & \underline{91.72}
& \underline{63.27} & \textbf{89.97}
& \textbf{63.69} & \textbf{81.09}
& \underline{50.88} & \textbf{75.76}
& \textbf{60.18} & 81.04
& 71.54 & 87.25
& \underline{66.27} & \textbf{87.38} \\

RSKT-SEG\cite{li2026exploring}%
$_{\textcolor{gray}{\text{\tiny\textit{AAAI2026}}}}$
& ViT-L & OVRSIS
& 69.43 & 93.69
& 55.63 & 92.37
& 74.15 & 90.27
& 63.57 & 87.12
& 57.05 & \underline{89.19}
& 51.28 & 78.35
& 38.94 & 70.26
& 48.87 & 74.70
& 34.78 & 68.26
& 54.86 & 82.69 \\

Pi-Seg
& ViT-L & OVRSIS
& \textbf{84.79} & \textbf{96.83}
& \textbf{58.70} & \textbf{93.01}
& \textbf{77.82} & \textbf{91.58}
& \textbf{77.99} & 90.15
& \textbf{66.90} & 88.72
& \underline{61.15} & \underline{80.91}
& \textbf{51.51} & 73.01
& \underline{60.17} & \underline{81.64}
& \textbf{79.90} & \textbf{88.84}
& \textbf{68.77} & 87.19 \\

\specialrule{0.95pt}{2pt}{2pt}
\end{tabular}
}

\vspace{-0.5em}
\caption{\textbf{Task transfer results on building extraction, road extraction, and flood detection.}}
\label{tab:03_main_table_other3task}
\end{table*}

\subsection{GlobalRSOV95K}
\label{sec:globalrsov95k}

GlobalRSOV95K is designed to provide a broader training foundation for OVRSIS and reduce dependence on any single source dataset. GlobalRSOV95K is constructed from three remote-sensing
segmentation resources: FAST, SIOR, and SOTA. It contains approximately \textbf{95K densely annotated images and 35 semantic categories} from five representative remote-sensing domains: \emph{town}, \emph{industrial}, \emph{forest}, \emph{waterfront}, and \emph{wasteland} (\cref{fig:pic_ovrsis95k}). These domains cover diverse geographic contexts, ranging from densely built environments and industrial facilities to vegetation-dominated regions, water-related scenes, and sparsely developed land. The resulting dataset exposes models to broader variations in scene composition and image appearance than an isolated training source.

The construction of GlobalRSOV95K consists of three stages. First, we collect remote-sensing images from diverse scenes within the five domains to improve geographic and visual coverage. Second, we consolidate the semantic concepts from different sources into a unified taxonomy, resolving inconsistent category names and definitions. Third, dense semantic masks are generated through a semi-automatic annotation pipeline and subsequently verified by human annotators. This process produces a large-scale training set with consistent category definitions and pixel-level supervision, enabling models to learn shared semantic correspondence from heterogeneous remote-sensing observations. Further details on domain selection, taxonomy construction, annotation, and quality control are provided in Appendix~\ref{sec:globalrsov95k_construction}.

\subsection{Unified Multi-Domain Evaluation}
\label{sec:unified_benchmark}

Based on GlobalRSOV95K, we establish a unified downstream evaluation suite for measuring transfer beyond the training distribution. The complete benchmark contains more than \textbf{170K annotated images and 122 semantic categories}, including the GlobalRSOV95K training set and 10 image-disjoint downstream OVRSIS datasets. These datasets span satellite, aerial, and unmanned aerial vehicle (UAV) imagery, thereby introducing substantial variation in spatial resolution and scene structure (\cref{fig:pic_fig_2_ovrsisv2}).

Unlike conventional single-source protocols, all methods in the unified setting are trained on the same GlobalRSOV95K foundation and directly evaluated on an identical collection of downstream datasets. This design reduces the influence of source-dataset selection and enables controlled comparison across methods. Importantly, no downstream evaluation image is included in GlobalRSOV95K, preventing image-level information leakage between training and testing.

As shown in \cref{fig:fig_test_centric_overlap_bar_GlobalRSOV95K}, the downstream taxonomies only partially overlap with the 35 training categories. The protocol therefore evaluates both transfer to categories observed during training and recognition of previously unseen categories. Beyond conventional semantic segmentation datasets, the benchmark also includes \textbf{building extraction, road extraction, and flood detection}, allowing us to examine whether open-vocabulary models can generalize to application-oriented geospatial tasks. The relation between GlobalRSOV95K and the downstream suite is further discussed in Appendix~\ref{sec:suppl_relation_unified_benchmark}.

\section{Experiments}
\label{sec:experiments}

\subsection{Experimental Setup}

\subsubsection{Single-source protocol.}
We first follow the established OVRSISBench settings, using DLRSD or iSAID as an individual training source and evaluating direct transfer to the corresponding downstream datasets. We retain the original category mappings and evaluation protocol to ensure comparability with prior work\cite{li2026exploring}.

\subsubsection{Unified multi-source protocol.}
All methods are then trained on \textbf{GlobalRSOV95K} and evaluated on the same image-disjoint downstream suite with unified category mappings. We use the prompt template $\textit{``a photo of \{class\}''}$. The input resolutions are $384\times384$ and $336\times336$ for ViT-B/16 and ViT-L/14, respectively. All trainable baselines are re-trained for 40K iterations on four NVIDIA H100 GPUs with a total batch size of 8, using AdamW, an initial learning rate of $2\times10^{-4}$, and cosine decay. We report mIoU and mACC for individual datasets, and m-mIoU and m-mACC averaged across datasets. Unless specified, $\sigma_t=0.02$ and $r=2$. Additional hyperparameter sensitivity and random-seed stability analyses are provided in Appendices~\ref{sec:suppl_hyperparameter_sensitivity} and~\ref{sec:suppl_seed_stability}.

\subsection{Main Results}

\subsubsection{Evaluation under Existing OVRSISBench Protocols}
\label{sec:single_source_results}

\cref{tab:01_main_table_DLRSD_iSAID} reports the results under the established single-source OVRSISBench protocols. Compared with its underlying CAT-Seg baseline, Pi-Seg improves m-mIoU/m-mACC by 8.68/10.49 and 6.75/9.38 points under DLRSD training with ViT-B and ViT-L, respectively, achieving the best mean performance in both settings. Under iSAID training, Pi-Seg also improves the baseline by 2.03/2.98 and 4.97/0.52 points. Although it does not outperform every specialized OVRSIS method in these settings, the consistent gains over the same baseline across both training sources and backbone scales indicate that the proposed perturbation mechanism is not specific to GlobalRSOV95K or a particular source distribution.

\subsubsection{From Single-Source to Unified Multi-Source Evaluation}
\label{sec:from_single_to_multi}

The single-source experiments evaluate transfer from individual training distributions. We next adopt a unified protocol in which all trainable methods use GlobalRSOV95K as the same training foundation and are evaluated on an identical image-disjoint downstream suite. This setting reduces the influence of source-dataset selection and enables a more controlled comparison of cross-domain transfer.

\subsubsection{Comparison on the Unified Multi-Source Benchmark}
\label{sec:unified_benchmark_results}

As shown in \cref{tab:02_main_table_OVRSIS95K}, Pi-Seg achieves the best mean performance under both backbone settings. With ViT-B, it surpasses the strongest competing results by 1.60 m-mIoU and 1.18 m-mACC; with ViT-L, the corresponding gains are 1.63 and 2.15 points. Pi-Seg does not dominate every individual dataset, but its superior average performance indicates more balanced transfer across heterogeneous downstream domains. Together with the single-source results, this demonstrates that its improvements persist under both source-specific and unified multi-source training protocols.

\subsubsection{Generalization to Application-Oriented Tasks}
\label{sec:downstream_tasks}

We further evaluate Pi-Seg on building extraction, road extraction, and flood detection. As reported in \cref{tab:03_main_table_other3task}, Pi-Seg achieves the best m-mIoU with both ViT-B and ViT-L, outperforming the strongest competing methods by 6.49 and 2.50 points, respectively. Its m-mACC remains competitive but is not consistently the highest. These results show that the main advantage of Pi-Seg lies in improving region-level overlap and spatial delineation across application-oriented geospatial tasks. Additional resolution-based robustness analysis is provided in Appendix~\ref{sec:suppl_resolution_robustness}.

\subsection{Ablation Study}

\subsubsection{Effect of Semantic Perturbation Modules}

\begin{table}[tb]
\centering
\footnotesize
\setlength{\tabcolsep}{4.5pt}
\vspace{-0.8em}
\resizebox{\linewidth}{!}{
\begin{tabular}{cccccc}
\toprule
Image Strategy & Text Strategy & Learnable & Backbone
& m-mIoU & m-mACC \\
\midrule
Prompt Tuning   & Prompt Tuning & \cmark & ViT-B & 34.72 & 51.21 \\
Random & Random & \xmark & ViT-B & 32.14 & 42.35 \\
\midrule
\xmark   & \xmark       & --     & ViT-B & 36.73 & 56.36 \\
Image-SPM & \xmark      & \cmark & ViT-B & 38.04 & 57.00 \\
\xmark   & Text-SPM     & \cmark & ViT-B & 36.41 & 54.21 \\
Image-SPM & Text-SPM    & \cmark & ViT-B & \bfseries 39.65
& \bfseries 57.63 \\
\bottomrule
\end{tabular}
}
\caption{
\textbf{Ablation of semantic perturbation and text adaptation
strategies.} Prompt tuning learns deterministic context.
}
\label{tab:ablation_SPM_tpami}
\vspace{-1em}
\end{table}

\cref{tab:ablation_SPM_tpami} verifies the effectiveness of the proposed semantic perturbation modules. Fixed random noise substantially
degrades performance, showing that unstructured feature corruption is harmful. Image-SPM alone improves the baseline, while Text-SPM alone is less effective without corresponding visual adaptation. Deterministic prompt tuning also underperforms the baseline, suggesting that simply introducing learnable prompts does not sufficiently address cross-domain appearance variations and may overfit the training distribution. In contrast, jointly applying Image-SPM and Text-SPM achieves the best performance, confirming the importance of coordinated, task-optimized perturbations across both modalities.
\subsubsection{Effect of Perturbation Distributions}

\begin{table}[t]
\centering
\scriptsize
\vspace{-0.6em}
\setlength{\tabcolsep}{4.0pt}
\renewcommand{\arraystretch}{1.05}
\resizebox{\linewidth}{!}{
\begin{tabular}{lcccc}
\toprule
\multirow{2}{*}{\textbf{Text perturb.}}
& \multicolumn{4}{c}{\textbf{Visual perturbation}} \\
\cmidrule(lr){2-5}
& Gaussian & Laplace & Student-$t$ & Uniform \\
\midrule
Gaussian
& 39.48 / 56.90
& 39.57 / 57.00
& 38.66 / 56.14
& 38.80 / 56.20 \\

Laplace
& \textbf{40.06} / 57.24
& 39.22 / 55.99
& 39.02 / \textbf{57.66}
& 38.91 / 56.25 \\

Student-$t$
& \underline{39.71} / 56.71
& 38.06 / 55.40
& 39.35 / 56.32
& 39.69 / \underline{57.39} \\

Uniform
& 39.69 / \underline{57.39}
& 38.68 / 56.91
& 39.23 / 55.89
& 39.67 / 57.27 \\
\bottomrule
\end{tabular}
}
\caption{\textbf{Effect of perturbation distributions.}}
\label{tab:noise_distribution_tpami}
\vspace{-3mm}
\end{table}

\cref{tab:noise_distribution_tpami} compares Gaussian, Laplace, Uniform, and Student-$t$ perturbations. Pi-Seg remains competitive across distribution combinations, indicating that its effectiveness does not depend on a specific noise family. The remaining variation reflects different feature-space exploration behaviors. A detailed analysis of Student-$t$ perturbation strength is provided in Appendix~\ref{sec:suppl_heavy_tailed}.

\subsection{Seen/Unseen Category Analysis}
\label{sec:seen_unseen}

\begin{table}[t]
    \centering
    \scriptsize
    \setlength{\tabcolsep}{2.5pt}
    \renewcommand{\arraystretch}{0.8}

    \resizebox{\columnwidth}{!}{%
    \begin{tabular}{llcccccc}
        \toprule
        \multirow{2}{*}{\textbf{Dataset}}
        & \multirow{2}{*}{\textbf{Method}}
        & \multicolumn{2}{c}{\textbf{Overall}}
        & \multicolumn{2}{c}{\textbf{Seen}}
        & \multicolumn{2}{c}{\textbf{Unseen}} \\
        \cmidrule(lr){3-4}
        \cmidrule(lr){5-6}
        \cmidrule(lr){7-8}
        & & mIoU & mACC & mIoU & mACC & mIoU & mACC \\
        \midrule

        \multirow{3}{*}{UAVid}
        & CAT-Seg
        & \underline{44.44} & \textbf{66.45}
        & \underline{65.17} & \textbf{85.36}
        & \underline{23.70} & \underline{47.53} \\

        & RSKT-Seg
        & 29.63 & 46.77
        & 46.77 & 75.77
        & 12.50 & 17.78 \\

        & Pi-Seg
        & \textbf{45.33} & \underline{64.26}
        & \textbf{66.37} & \underline{78.70}
        & \textbf{24.30} & \textbf{49.83} \\

        \midrule

        \multirow{3}{*}{UDD5}
        & CAT-Seg
        & \underline{56.33} & \underline{81.29}
        & \underline{53.71} & \underline{82.93}
        & \underline{64.17} & \underline{76.35} \\

        & RSKT-Seg
        & 34.48 & 54.86
        & 34.77 & 61.02
        & 33.62 & 36.40 \\

        & Pi-Seg
        & \textbf{68.45} & \textbf{88.29}
        & \textbf{65.18} & \textbf{86.81}
        & \textbf{78.26} & \textbf{92.73} \\

        \bottomrule
    \end{tabular}%
    }

    \caption{\textbf{Seen/unseen category analysis.}}
    \label{tab:seen_unseen}
    \vspace{-1em}
\end{table}

As shown in \cref{tab:seen_unseen}, Pi-Seg achieves the highest seen and
unseen mIoU on both datasets. Its clear gains on unseen categories indicate
that semantic perturbation improves open-vocabulary transfer rather than
merely fitting the seen classes.

\subsection{Efficiency Comparison}
\label{sec:efficiency}

\begin{table}[t]
    \centering
    \scriptsize
    \setlength{\tabcolsep}{2.3pt}
    \renewcommand{\arraystretch}{0.8}

    \begin{tabular*}{\columnwidth}{
        @{\extracolsep{\fill}}
        llccccc
        @{}
    }
        \toprule
        \textbf{Backbone}
        & \textbf{Method}
        & \makecell{\textbf{Params}\textbf{(M)}}
        & \makecell{\textbf{GFLOPs}$\textbf{(t/e)}$}
        & \makecell{\textbf{Memory}\textbf{(GB)}}
        & \textbf{FPS} \\
        \midrule

        \multirow{4}{*}{ViT-B}
        & CAT-Seg
        & 154.29 & 608.62 / 608.62
        & 14.10 & 66.30 \\

        & RSKT-Seg
        & 398.07 & 572.65 / 572.65
        & 15.53 & 64.84 \\

        & RSKT-Seg$^{*}$
        & 398.07 & 572.65 / 3014.82
        & 15.53 & 24.19 \\

        & Pi-Seg
        & 155.60 & 724.20 / 608.62
        & 14.13 & 66.89 \\

        \midrule

        \multirow{4}{*}{ViT-L}
        & CAT-Seg
        & 433.73 & 1249.48 / 1249.48
        & 20.54 & 56.70 \\

        & RSKT-Seg
        & 678.29 & 1125.64 / 1125.64
        & 19.31 & 64.27 \\

        & RSKT-Seg$^{*}$
        & 678.29 & 1125.64 / 5798.10
        & 19.31 & 19.51 \\

        & Pi-Seg
        & 436.68 & 1417.59 / 1249.48
        & 20.59 & 57.92 \\

        \bottomrule
    \end{tabular*}

    \caption{\textbf{Efficiency comparison.}
    GFLOPs are reported as training/evaluation costs.
    RSKT-Seg$^{*}$ uses sliding-window inference.
    All methods use the same hardware, batch size, and $600\times800$ input.}
    \label{tab:efficiency}
    \vspace{-3mm}
\end{table}

As shown in \cref{tab:efficiency}, Pi-Seg adds few parameters over CAT-Seg.
Its perturbation modules increase training cost but introduce no inference
overhead, while avoiding the high cost of sliding-window inference.

\subsection{Perturbation Dynamics of Pi-Seg}
\label{sec:delta_logs}

\begin{figure}[tb]
    \centering
    \includegraphics[width=\linewidth]{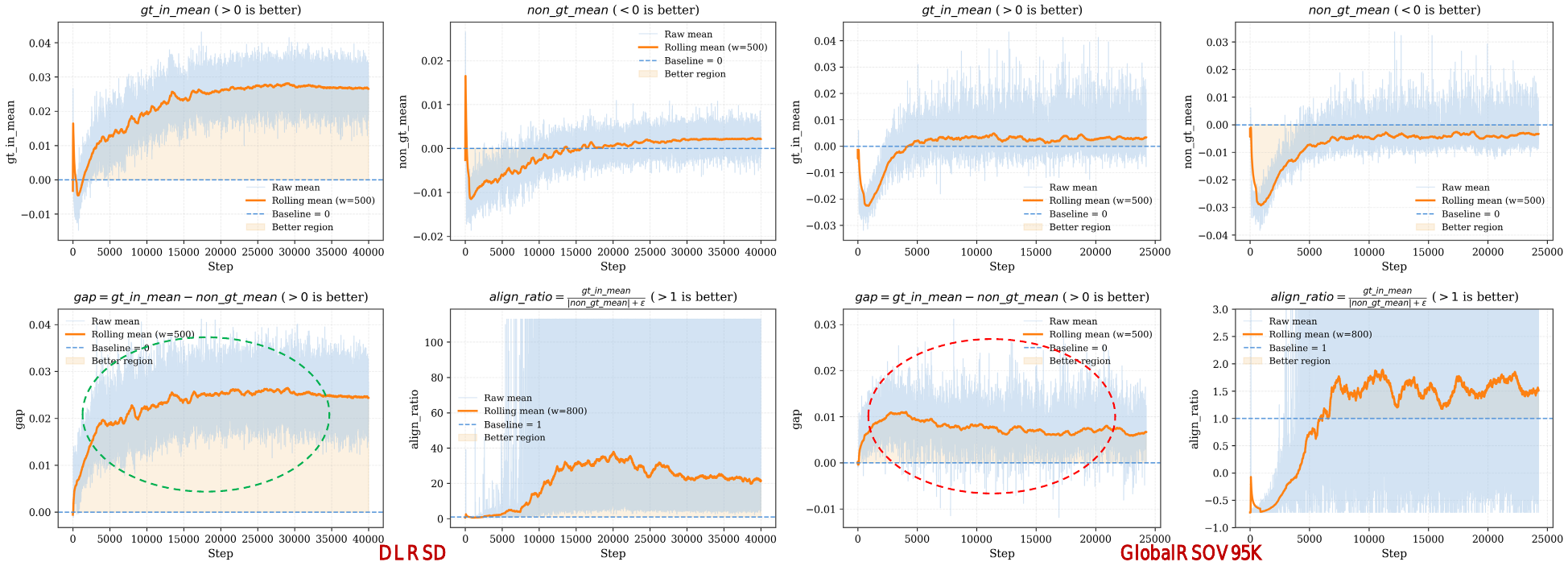}
    \caption{\textbf{Training dynamics of Pi-Seg.}}
    \label{fig:pic_delta_logs}
    \vspace{-1mm}
\end{figure}

We analyze four $\Delta$-correlation statistics in \cref{fig:pic_delta_logs}. \textit{gt\_in\_mean} and \textit{non\_gt\_mean} measure correlation changes inside and outside ground-truth regions, while \textit{gap} and \textit{align\_ratio} measure their separation and relative target alignment. Across training distributions, target-region correlation becomes positive after warm-up, whereas non-target correlation remains near or below zero. The increasing separation shows that Pi-Seg learns selective, task-favorable variations rather than unstructured random noise. Formal definitions of these statistics are provided in Appendix~\ref{sec:suppl_delta_corr_stats}.
The trajectories under GlobalRSOV95K are more conservative than those under DLRSD, reflecting the greater diversity of multi-source training. Nevertheless, Pi-Seg consistently maintains target enhancement and non-target suppression, showing that the same perturbation principle remains effective from single-source to multi-source settings.

\subsection{Qualitative and Cost Map Analysis}
\label{sec:qualitative_analysis}

\begin{figure}[tb]
    \centering

    \begin{minipage}{\linewidth}
        \centering
        \includegraphics[width=\linewidth]{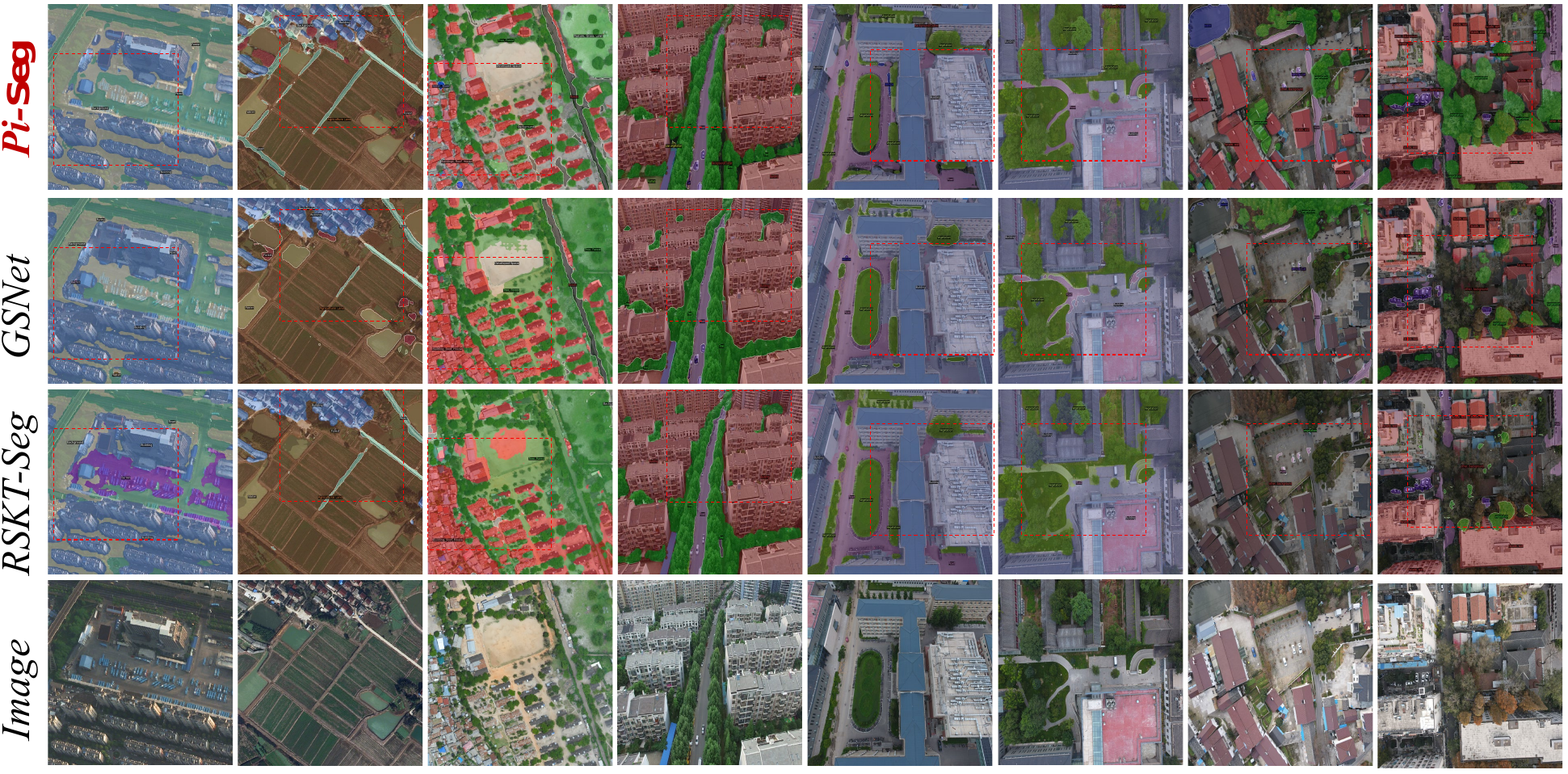}
        \small \textbf{(a)} General OVRSIS datasets.
    \end{minipage}

    \vspace{0.4em}

    \begin{minipage}{\linewidth}
        \centering
        \includegraphics[width=\linewidth]{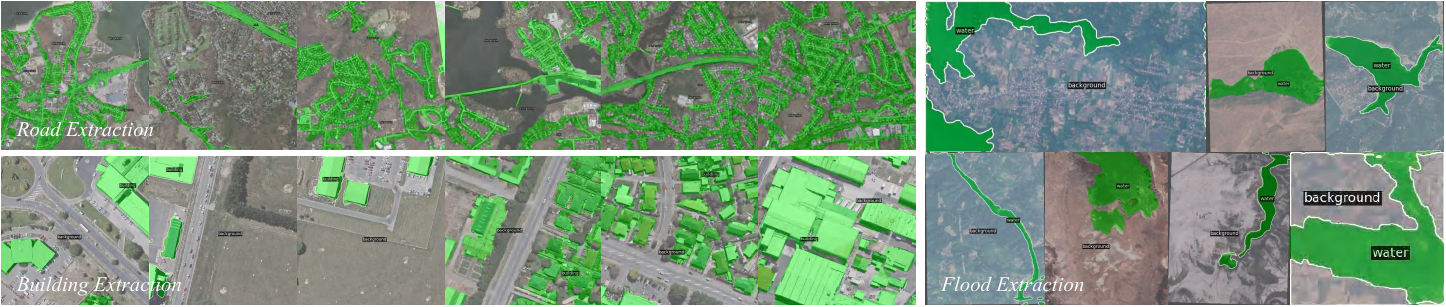}
        \vspace{-0.8em}

        \small \textbf{(b)} Application-oriented tasks.
    \end{minipage}

    \caption{\textbf{Qualitative segmentation results.}}
    \label{fig:qualitative_results}
    \vspace{-0.8em}
\end{figure}

\begin{figure}[t]
    \centering

    \begin{minipage}{\linewidth}
        \centering
        \includegraphics[width=\linewidth]{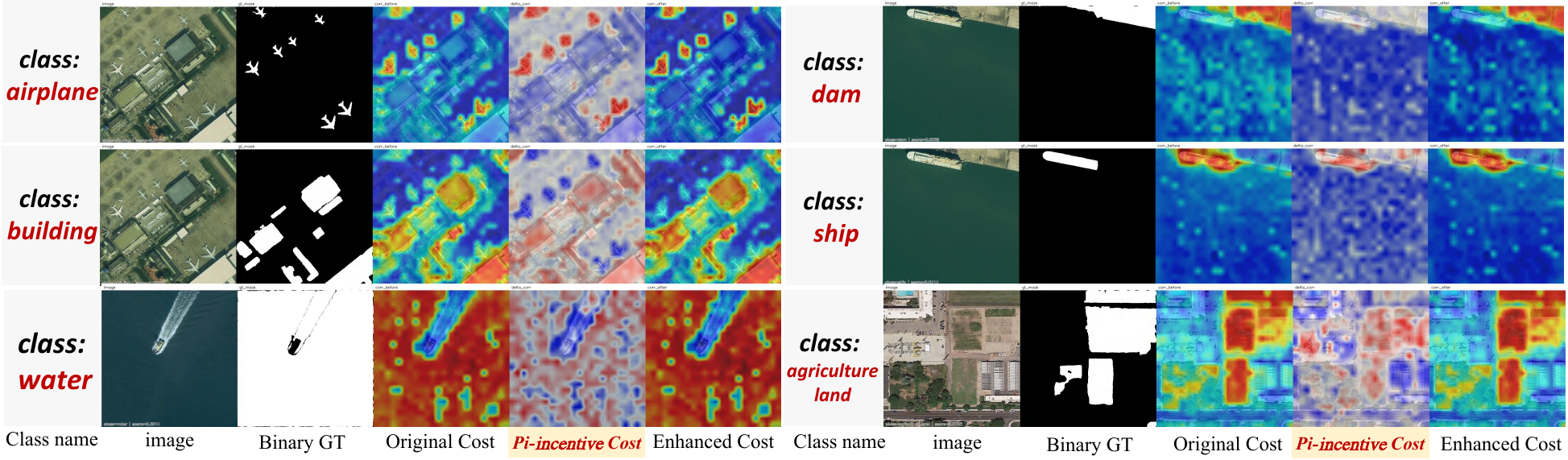}
        \vspace{-1em}
        
        \small \textbf{(a)} Fine-grained cost refinement.
    \end{minipage}

    \vspace{0.4em}

    \begin{minipage}{0.95\linewidth}
        \centering
        \includegraphics[width=\linewidth]{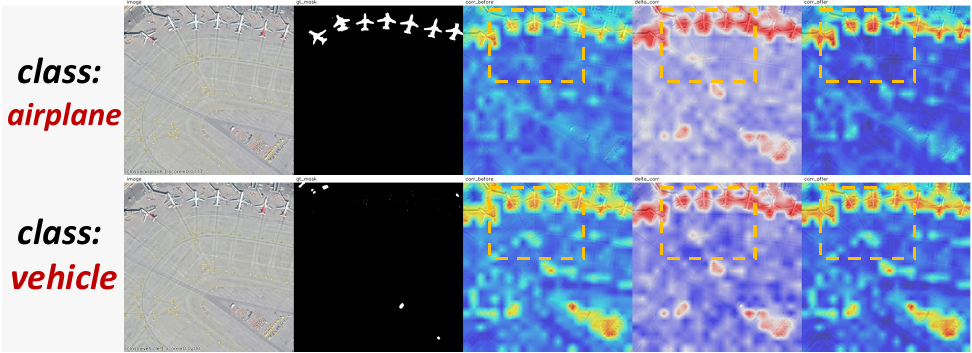}
        \vspace{-0.6em}
        \small \textbf{(b)} Representative failure cases.
    \end{minipage}
    \caption{\textbf{Cost feature map analysis of Pi-Seg.}}
    \label{fig:pic_cost_and_failure}
    \vspace{-0.8em}
\end{figure}

\cref{fig:qualitative_results} shows that Pi-Seg produces more complete target regions and suppresses spurious background responses across general OVRSIS datasets and application-oriented tasks. The cost maps in \cref{fig:pic_cost_and_failure} (a)further reveal more localized category responses after perturbation learning.

\subsection{Failure Case Analysis}
\cref{fig:pic_cost_and_failure} (b) show that Pi-Seg still confuse fine-grained categories such as \emph{ship} and \emph{airplane} with the broader category \emph{vehicle}. This limitation mainly stems from coarse semantic prototypes derived from short class prompts and motivates richer textual descriptions in future work.

\section{Conclusion}

We presented \textbf{Pi-Seg}, a lightweight perturbation-based framework that improves pixel--text alignment through learnable variations of textual prototypes and dense visual features. Its consistent gains under the established OVRSISBench protocols demonstrate robustness.
We also introduced \textbf{GlobalRSOV95K}, containing 95K images and 35 categories, together with a unified multi-source benchmark covering over 170K images, 122 categories, and 10 downstream datasets. Results under both single-source and multi-source protocols show that Pi-Seg improves cross-dataset and cross-task transfer without additional inference-time computation.

\section*{Acknowledgments}

This work was supported in part by the Natural Science Foundation of China
under Grant 62306241, and in part by grants from the Innovation Foundation
for Doctor Dissertation of Northwestern Polytechnical University
(No.~CX2025109).

Bingyu Li conducted this work during an internship at the Institute of
Artificial Intelligence (TeleAI), China Telecom.

\bibliography{aaai2027}

\clearpage
\appendix
\setcounter{section}{0}
\setcounter{secnumdepth}{3}
\renewcommand{\thesection}{A.\arabic{section}}
\renewcommand{\thesubsection}{\thesection.\arabic{subsection}}
\renewcommand{\thesubsubsection}{\thesubsection.\arabic{subsubsection}}

\section{Supplementary Overview}
\label{sec:suppl_overview}

This supplementary material provides additional details and analyses that complement the main paper. 
We first present the construction details of \textbf{GlobalRSOV95K}, including its data sources, training--evaluation isolation, domain coverage, taxonomy design, semi-automatic annotation pipeline, and quantitative quality-control procedure. 
We then review the cost-aggregation formulation used by dense open-vocabulary segmentation models and clarify where Pi-Seg introduces semantic perturbation. 
After that, we report additional results under the earlier single-source OVRSISBench protocols and provide extended analyses on hyperparameter sensitivity, random-seed stability, seen/unseen category transfer, heavy-tailed perturbation, and resolution robustness. 
Finally, we discuss representative failure cases and limitations.

These supplementary analyses further support the main findings of the paper. 
They show that GlobalRSOV95K provides a diverse and reliable training foundation, and that Pi-Seg remains effective under different training protocols, stochastic settings, category splits, perturbation distributions, and image resolutions.

For convenient reference, this appendix is numbered as follows. 
\cref{sec:globalrsov95k_construction} presents dataset construction details; 
\cref{sec:preliminaries} reviews the OVS and cost-aggregation preliminaries; 
\cref{sec:suppl_robustness} provides robustness and ablation analyses; 
\cref{sec:suppl_failure_cases} discusses failure cases and limitations; 
and \cref{sec:suppl_variational_piseg,sec:open_vocabulary_generalization} provide the variational and open-vocabulary generalization interpretations of Pi-Seg.

\section{Construction Details of GlobalRSOV95K}
\label{sec:globalrsov95k_construction}

\subsection{Dataset Scope and Design Principle}
\label{sec:suppl_dataset_scope}

GlobalRSOV95K is designed as a shared multi-domain training foundation for open-vocabulary remote sensing image segmentation. It contains approximately 95K annotated images and 35 semantic categories across five representative remote-sensing domains: \emph{town}, \emph{industrial}, \emph{forest}, \emph{waterfront}, and \emph{wasteland}. 
The design goal of GlobalRSOV95K is not to simply increase the number of images. Instead, the dataset is constructed to provide a balanced and reusable training foundation for OVRSIS. Existing remote-sensing segmentation datasets often focus on a particular sensor, city, resolution, or application scenario. Such datasets are valuable for supervised segmentation, but they are less suitable as a general training source for open-vocabulary transfer because their categories and scene distributions are usually narrow. GlobalRSOV95K therefore emphasizes three properties: broad domain coverage, a unified semantic taxonomy, and scalable annotation. Together, these properties allow different OVRSIS models to be trained under the same foundation and compared under a fixed downstream suite.

The five domains are selected to cover both natural and built environments. \emph{Town} scenes contain dense residential layouts, fragmented objects, and complex road-building interactions. \emph{Industrial} scenes contain large artificial structures and regular geometric patterns. \emph{Forest} scenes are dominated by vegetation textures and natural boundaries. \emph{Waterfront} scenes contain land-water interfaces and objects such as boats, docks, and coastal structures. \emph{Wasteland} scenes include sparse vegetation, bare soil, and open terrain. This domain composition helps expose models to heterogeneous visual structures during training, which is important for downstream transfer.

\subsection{Data Sources and Training--Evaluation Isolation}
\label{sec:suppl_data_sources_isolation}

GlobalRSOV95K is constructed from three remote-sensing segmentation resources: \textbf{FAST}, \textbf{SIOR}, and \textbf{SOTA}. These resources are used exclusively to form the training foundation. During benchmark construction, we explicitly separate the training-image pool from all downstream evaluation datasets. No image contained in any downstream evaluation set is included in GlobalRSOV95K, and no downstream evaluation image is reused during taxonomy construction, mask generation, or model training. Consequently, all results in the unified benchmark are obtained under an image-disjoint training--evaluation protocol.

It is important to distinguish \emph{image overlap} from \emph{category overlap}. Some downstream datasets share category names or semantically related concepts with the 35-class training taxonomy; this semantic overlap is intentionally retained to evaluate seen-category transfer. Other downstream categories are absent from the training taxonomy and are used to evaluate open-vocabulary recognition. However, category overlap does not imply sample reuse: the images used to construct GlobalRSOV95K and those used for downstream evaluation are mutually exclusive. This separation prevents direct image memorization and ensures that the reported performance measures cross-dataset transfer rather than repeated evaluation on training images.

\subsection{Taxonomy Construction}
\label{sec:suppl_taxonomy}

A major challenge in constructing an OVRSIS training foundation is category consistency. Different remote sensing datasets may use different names for similar semantic concepts, or may annotate the same region at different semantic granularities. For example, similar built-up regions may be labeled as \emph{building}, \emph{house}, \emph{residential area}, or \emph{developed space}; vegetation-related regions may appear as \emph{tree}, \emph{grass}, \emph{low vegetation}, or \emph{forest}. Directly merging such labels would introduce ambiguity into text--mask supervision.

To avoid this problem, we organize the dataset with a unified taxonomy. Candidate category names are cleaned by removing redundant synonyms, merging visually indistinguishable labels, and keeping categories that have sufficient visual evidence in the image domain. This taxonomy cleaning step is especially important for open-vocabulary training, because the textual category name directly serves as the semantic query. If two names describe nearly identical visual regions, the model may receive inconsistent supervision. Conversely, if a category is too broad, it may weaken fine-grained segmentation. The final 35 categories are selected to balance semantic diversity and annotation reliability.

\subsection{Annotation Pipeline}
\label{sec:suppl_annotation_pipeline}

\begin{figure}[tb]
    \centering
    \includegraphics[width=\linewidth]{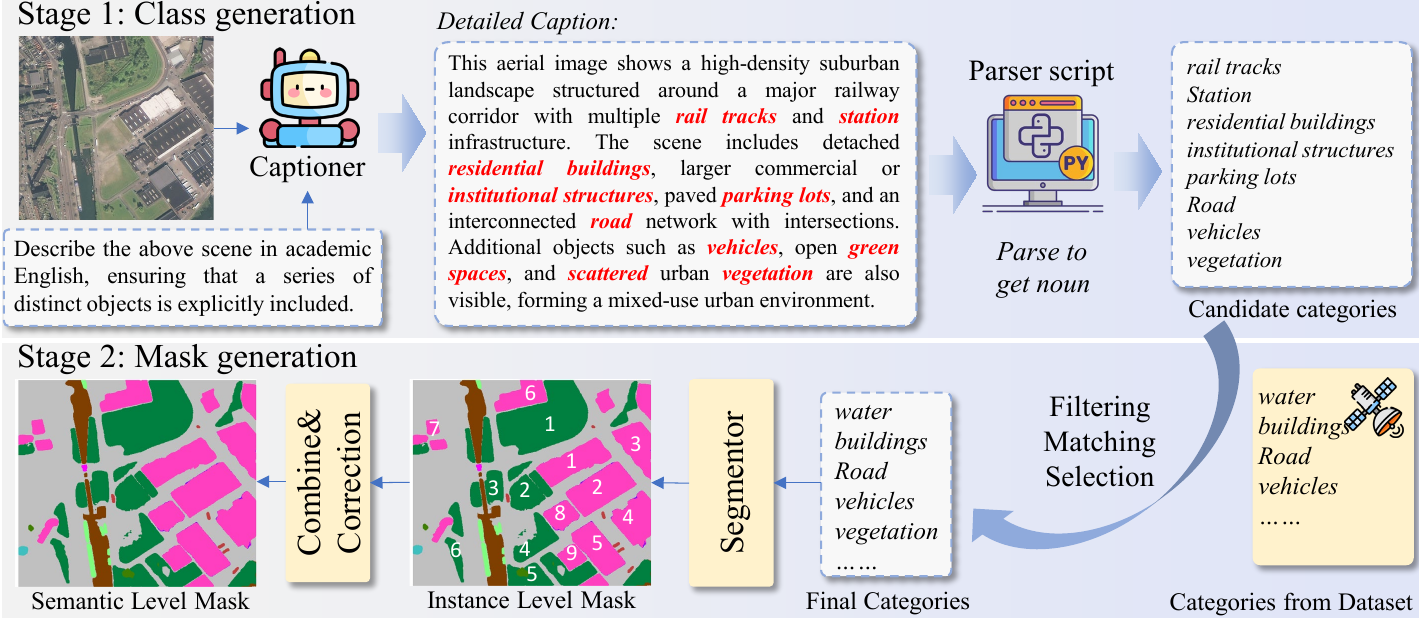}
    \caption{\textbf{Construction pipeline of GlobalRSOV95K.} Candidate categories are derived from scene descriptions, followed by mask generation and human verification.}
    \label{fig:pic_fig_2_how2getting_dataset}
    \vspace{-0.5em}
\end{figure}

The construction pipeline is shown in \cref{fig:pic_fig_2_how2getting_dataset}. It follows a semi-automated design that combines caption-guided category discovery, mask proposal generation, and human verification. First, candidate semantic concepts are extracted from scene-level descriptions and aligned with the predefined dataset taxonomy. This step helps identify which categories are likely to appear in a given image and reduces unnecessary mask generation for absent classes. Second, mask proposals are generated for the selected categories and converted into pixel-level semantic annotations. Third, annotators verify the generated masks, remove unreliable regions, and correct category assignments when necessary. This procedure improves annotation efficiency while maintaining semantic consistency across domains.

The pipeline is designed to reduce annotation cost without fully relying on automatic pseudo-labels. In remote-sensing imagery, objects may be small, densely arranged, or partially occluded, and scene categories can have fuzzy boundaries. Fully automatic mask generation may therefore produce incomplete object regions or confuse visually similar classes. Human verification remains necessary to ensure that training masks provide reliable supervision for dense pixel--text alignment. In practice, annotators focus on correcting category mismatches, removing masks with severe boundary errors, and filtering proposals that correspond to background clutter.

\subsection{Human Audit and Quality Control}
\label{sec:suppl_quality_control}

Although GlobalRSOV95K is constructed using a scalable semi-automatic pipeline, we further conduct a large-scale human audit to quantitatively evaluate its annotation quality. Specifically, we randomly sample \textbf{20,000} images covering all five scene domains and inspect all generated category proposals and their associated semantic masks. The category-related ratios below are computed over the audited category proposals, whereas the mask-related ratios are computed over the audited semantic masks. \cref{tab:human_audit} summarizes the resulting statistics.

The audit is conducted at two levels. First, the \textbf{category audit} checks whether each generated category is present in the image and semantically consistent with the unified taxonomy after category matching. Second, the \textbf{mask audit} checks whether the corresponding semantic mask is aligned with the correct object or region, including its spatial extent and boundaries.

At the category level, \textbf{97.25\%} of the automatically generated category proposals are accepted without modification, \textbf{2.22\%} require manual relabeling or taxonomy correction, and \textbf{0.53\%} are discarded as invalid or false-positive proposals. At the mask level, \textbf{91.66\%} of the masks are accepted without spatial modification, while the remaining \textbf{8.34\%} require boundary refinement, region merging, or filtering of incorrectly included instances. The discarded category ratio therefore corresponds to a false-positive rate of approximately \textbf{0.53\%} before manual correction, mainly caused by visually confusing categories or over-segmentation in cluttered scenes.

All errors identified in the audited subset are corrected or removed before the final release. The same unified taxonomy and quality criteria are applied across the five domains so that a category retains consistent semantics despite changes in scale, appearance, and surrounding context. For example, \emph{road} regions in town and forest scenes may differ in width, material, and neighboring objects, but they are audited according to the same semantic definition. These procedures reduce category ambiguity and mask noise, thereby improving the reliability of GlobalRSOV95K as a training foundation for dense pixel--text alignment.

\begin{table}[t]
\centering
\resizebox{\linewidth}{!}{
\begin{tabular}{lcc}
\toprule
Item & Count / Ratio & Description \\
\midrule
Audited images & 20,000 & Images included in the human-audit subset \\
Audited scene domains & 5 & Town / industrial / forest / waterfront / wasteland \\
\midrule
Category accepted & 97.25\% & Correct categories retained without modification \\
Category corrected & 2.22\% & Categories manually corrected after taxonomy review \\
Category discarded & 0.53\% & Invalid or false-positive category proposals removed \\
\midrule
Mask accepted & 91.66\% & Masks retained without spatial modification \\
Mask corrected & 8.34\% & Masks refined by boundary editing, merging, or filtering \\
\bottomrule
\end{tabular}}
\caption{Human-audit statistics of the GlobalRSOV95K annotation pipeline. The audited images cover all five scene domains. Category ratios are computed over audited category proposals, while mask ratios are computed over audited semantic masks.}
\label{tab:human_audit}
\vspace{-1em}
\end{table}

\subsection{Comparison with Existing Remote-Sensing Datasets}
\label{sec:suppl_dataset_comparison}

\begin{table*}[t]
    \centering
    \renewcommand{\arraystretch}{0.95}
    \setlength{\tabcolsep}{1.0pt}
    \resizebox{\linewidth}{!}{
    \begin{tabular}{lllllllll}
        \toprule
        \textbf{Dataset / Benchmark}
        & \textbf{Year}
        & \textbf{Images}
        & \textbf{Categories}
        & \textbf{Task(s)}
        & \textbf{GSD (m)}
        & \textbf{Image Width}
        & \textbf{Total Area ($\mathrm{km}^2$)}
        & \textbf{Source} \\
        \midrule

        LandCover.ai
        & 2020 & 41 & 3
        & Building/Road Seg.
        & 0.25--0.5 & 800--13,000 & 216.27
        & \cite{boguszewski2021landcover} \\

        LoveDA
        & 2021 & 5,987 & 7
        & Semantic Seg.
        & 0.3 & 1,024 & 536.15
        & \cite{wang2021loveda} \\

        SOTA
        & 2023 & 17,480 & 18
        & Semantic Seg.
        & -- & 1,024 & --
        & \cite{wang2023samrs} \\

        LandDiscover50K
        & 2024 & 51,846 & 40
        & OVRSIS
        & -- & 400--1,200 & --
        & \cite{ye2025towards} \\

        OVRSISBench (DLRSD)
        & 2025 & 55,628 & 72
        & OVRSIS
        & -- & 256--4,000 & --
        & \cite{li2026exploring} \\

        OVRSISBench (iSAID)
        & 2025 & 50,666 & 72
        & OVRSIS
        & -- & 256--4,000 & --
        & \cite{li2026exploring} \\

        \midrule

        \textbf{Ours: Unified OVRSIS Benchmark}
        & 2026 & \textbf{170,036} & \textbf{122}
        & \textbf{OVRSIS + Task Transfer}
        & -- & \textbf{256--4,000} & -- & -- \\

        \quad GlobalRSOV95K training foundation
        & 2026 & 94,620 & 35
        & OVRSIS training
        & -- & 256--4,000 & -- & -- \\

        \quad Fixed downstream evaluation suite
        & 2026 & 75,416 & 88
        & Cross-dataset evaluation
        & -- & 256--4,000 & -- & -- \\

        \bottomrule
    \end{tabular}}
    \caption{\textbf{Comparison with existing remote-sensing datasets and OVRSIS benchmarks.}
    GlobalRSOV95K denotes the common multi-domain training foundation, whereas the unified OVRSIS benchmark additionally includes a fixed image-disjoint downstream evaluation suite.}
    \label{tab:dataset_comparison}
\end{table*}

As summarized in \cref{tab:dataset_comparison}, existing remote-sensing segmentation datasets are typically designed for predefined semantic categories or specific extraction tasks. Recent OVRSIS datasets broaden the category space, but their evaluations generally rely on individual training sources and source-dependent downstream configurations. In contrast, our benchmark separates a common training foundation from a fixed image-disjoint evaluation suite. GlobalRSOV95K provides 94,620 training images from multiple remote-sensing domains, while the downstream suite contains 75,416 images that are excluded from training.
The complete benchmark covers 170,036 images and 122 semantic categories, making it substantially larger than the individual datasets and source-specific OVRSISBench settings listed in the table. Moreover, it evaluates not only cross-dataset semantic segmentation but also transfer to building extraction, road extraction, and flood detection. This design enables controlled comparison under a shared training distribution while retaining substantial variation in image resolution, sensing platform, geographic context, and downstream taxonomy.

\subsection{Relation to the Unified Benchmark}
\label{sec:suppl_relation_unified_benchmark}

GlobalRSOV95K serves as the training foundation, while the downstream benchmark evaluates generalization. This separation is central to the proposed protocol. The training set provides diverse supervision, but the downstream suite contains partially disjoint categories and different scene distributions. As a result, evaluation does not simply measure how well a model fits the training distribution; it also measures whether the learned dense vision--language alignment transfers to new datasets and geospatial tasks.

This design differs from isolated source--target transfer settings, where each paper may choose a different training source and a different target dataset. By fixing the training foundation and downstream evaluation suite, the proposed benchmark reduces protocol ambiguity and makes model comparison more meaningful.

\section{Preliminaries}
\label{sec:preliminaries}

\subsection{Open-Vocabulary Segmentation}
\label{sec:suppl_ovs_prelim}

Open-vocabulary semantic segmentation (OVS) predicts pixel-level semantic regions from an arbitrary category set, where test categories may not appear during training. Given an image $I \in \mathbb{R}^{H \times W \times 3}$ and category texts $\mathcal{C}=\{t_1,t_2,\ldots,t_N\}$, the goal is to predict a segmentation map $S \in \{1,\ldots,N\}^{H \times W}$. Unlike closed-set segmentation, the inference category set $\mathcal{C}_{test}$ may differ from the training category set $\mathcal{C}_{train}$.

Vision--language models such as CLIP~\cite{radford2021learning} provide a shared embedding space for images and texts. Let $\Phi^V$ and $\Phi^L$ denote the image and text encoders. They extract dense visual features $V \in \mathbb{R}^{H' \times W' \times D}$ and normalized text embeddings $T \in \mathbb{R}^{N \times D}$, respectively. OVS models estimate the semantic likelihood of each spatial location by matching dense visual features with category embeddings.

In the remote-sensing setting, this formulation must handle stronger visual variation than many natural-image scenes. The same textual category can correspond to different object scales, shapes, and contextual layouts. For example, a \emph{building} may appear as a small detached house, a large industrial warehouse, or a dense urban block. This makes dense matching between visual features and textual prototypes more fragile. The motivation of Pi-Seg is to improve this dense matching process by regularizing the local feature neighborhood during training.

\subsection{Cost Aggregation}
\label{sec:suppl_cost_aggregation}

Following cost-aggregation-based OVS~\cite{cho2024cat}, a multimodal cost volume $C \in \mathbb{R}^{H' \times W' \times N}$ is constructed using dense cosine similarity:
\begin{equation}
    C(i,n)=
    \frac{V(i)\cdot T(n)}
    {\|V(i)\|_2\|T(n)\|_2},
    \label{eq:suppl_cost_volume}
\end{equation}
where $V(i)$ and $T(n)$ denote the visual feature at position $i$ and the text embedding of class $n$, respectively. Since raw CLIP similarities are often noisy for dense prediction, a learnable aggregation module refines the cost volume:
\begin{equation}
    \tilde{C}=\mathcal{F}_{\mathrm{agg}}(C),
    \label{eq:suppl_cost_aggregation}
\end{equation}
where $\mathcal{F}_{\mathrm{agg}}$ models spatial context and inter-class relations to improve local consistency and category discrimination. Pi-Seg builds upon this formulation and regularizes the feature space before cost construction.

The cost volume can be interpreted as a dense semantic matching field. Each spatial location is compared with all category embeddings, and the resulting responses are later converted into segmentation logits. If the visual and textual features are poorly aligned, the cost volume may contain diffuse activations, spurious background responses, or incomplete target regions. Cost aggregation can improve local coherence, but it still depends on the quality of the underlying feature correspondence. Pi-Seg therefore perturbs text and visual features before cost construction, encouraging the model to maintain stable semantic correspondence under feature variations.

\section{Additional Robustness and Ablation Studies}
\label{sec:suppl_robustness}

\subsection{Hyperparameter Sensitivity}
\label{sec:suppl_hyperparameter_sensitivity}

\begin{table}[tb]
\centering
\footnotesize
\setlength{\tabcolsep}{5pt}
\resizebox{\linewidth}{!}{
\begin{tabular}{cccccc}
\toprule
\multicolumn{3}{c}{(a) Feature reduction ratio ($\sigma_t=0.02$)} &
\multicolumn{3}{c}{(b) Text noise std ($r=2$)} \\
\cmidrule(lr){1-3} \cmidrule(lr){4-6}
Reduction & m-mIoU & m-mACC & Std & m-mIoU & m-mACC \\
\midrule
1 & 37.25 & 55.60 & 0     & 38.39 & 55.34 \\
\textbf{2} & \textbf{39.23} & \textbf{57.25} & 0.005 & 37.46 & 55.93 \\
4 & 38.78 & 56.79 & 0.01  & 38.61 & 56.44 \\
8 & \underline{38.91} & \underline{57.21} & \textbf{0.02} & \textbf{38.91} & \underline{56.72} \\
- & - & - & 0.05  & \underline{38.63} & \textbf{57.69} \\
\bottomrule
\end{tabular}
}
\caption{Sensitivity analysis of key hyperparameters in Pi-Seg with ViT-B.}
\label{tab:sensitivity_hparams_tpami}
\vspace{-0.5em}
\end{table}

\cref{tab:sensitivity_hparams_tpami} evaluates two key hyperparameters of Pi-Seg. For Image-SPM, $r=2$ achieves the best overall performance, indicating that moderate feature compression balances representation capacity and optimization stability. A very small reduction ratio preserves more feature dimensions, but it may increase optimization complexity. A very large reduction ratio improves efficiency but may weaken the ability of Image-SPM to predict expressive spatial perturbations. The best performance at $r=2$ suggests that Image-SPM benefits from compact yet sufficiently informative semantic conditioning.

For Text-SPM, $\sigma_t=0.02$ yields the best m-mIoU, while $\sigma_t=0.05$ achieves the highest m-mACC. This behavior indicates that the perturbation strength should be large enough to expand the semantic neighborhood, but not so large that it disrupts the original CLIP text prototype. Overall, Pi-Seg is stable within moderate parameter ranges and benefits from controlled perturbation strength.

\subsection{Random-Seed Stability}
\label{sec:suppl_seed_stability}

\begin{table}[tb]
\centering
\footnotesize
\setlength{\tabcolsep}{8pt}
\resizebox{0.85\linewidth}{!}{
\begin{tabular}{lcc}
\toprule
Backbone & m-mIoU & m-mACC \\
\midrule
Baseline (ViT-B) & 36.73 & 56.36 \\
\midrule
Pi-Seg (ViT-B) & 39.25 $\pm$ 0.56 & 57.53 $\pm$ 0.29 \\
Pi-Seg (ViT-L) & 44.24 $\pm$ 0.48 & 63.01 $\pm$ 0.28 \\
\bottomrule
\end{tabular}
}
\caption{Pi-Seg performance statistics over 10 runs with different random seeds. The baseline row is the single-run reference reported in the main comparison.}
\label{tab:seed_mean_std}
\vspace{-0.5em}
\end{table}

We repeat Pi-Seg training 10 times with different random seeds. As shown in \cref{tab:seed_mean_std}, Pi-Seg consistently exceeds the single-run baseline reference and exhibits small standard deviations under both ViT-B and ViT-L. These results indicate that the reported improvements are stable across random initialization and stochastic sampling.

This analysis is particularly relevant because Pi-Seg explicitly introduces stochastic perturbations during training. A potential concern is that the performance gain may depend on a favorable random seed. The small standard deviations alleviate this concern and show that semantic perturbation provides stable regularization rather than unstable random augmentation. The ViT-L results show similarly low variance, indicating that the mechanism remains reliable when scaling the backbone.

\subsection{Seen/Unseen Category Analysis}
\label{sec:suppl_seen_unseen}

\begin{table*}[tb]
\centering
\small
\resizebox{\linewidth}{!}{
\begin{tabular}{lcccccccccccccccccc}
\toprule
\multirow{2}{*}{Method}
& \multicolumn{2}{c}{Building}
& \multicolumn{2}{c}{Road}
& \multicolumn{2}{c}{Tree}
& \multicolumn{2}{c}{Low vegetation}
& \multicolumn{2}{c}{Car}
& \multicolumn{2}{c}{Human}
& \multicolumn{2}{c}{Overall}
& \multicolumn{2}{c}{Seen}
& \multicolumn{2}{c}{Unseen} \\
\cmidrule(lr){2-3} \cmidrule(lr){4-5} \cmidrule(lr){6-7} \cmidrule(lr){8-9}
\cmidrule(lr){10-11} \cmidrule(lr){12-13} \cmidrule(lr){14-15} \cmidrule(lr){16-17} \cmidrule(lr){18-19}
& IoU & ACC & IoU & ACC & IoU & ACC & IoU & ACC & IoU & ACC & IoU & ACC & mIoU & mACC & mIoU & mACC & mIoU & mACC \\
\midrule
CAT-Seg  & \underline{82.72} & \textbf{95.53} & \underline{54.31} & \underline{90.34} & \textbf{58.49} & \textbf{70.21} & \textbf{46.50} & \textbf{71.58} & \underline{23.52} & \underline{64.17} & \textbf{1.09} & \underline{6.84} & \underline{44.44} & \textbf{66.45} & \underline{65.17} & \textbf{85.36} & \underline{23.70} & \underline{47.53} \\
RSKT-Seg & 73.07 & 89.26 & 27.83 & \textbf{94.68} & 39.41 & 43.36 & 36.07 & 51.87 & 1.42 & 1.48 & 0.00 & 0.00 & 29.63 & 46.77 & 46.77 & 75.77 & 12.50 & 17.78 \\
Pi-Seg   & \textbf{84.29} & \underline{90.29} & \textbf{60.36} & 85.85 & \underline{54.46} & \underline{59.95} & \underline{45.62} & \underline{67.09} & \textbf{26.21} & \textbf{70.34} & \underline{1.06} & \textbf{12.06} & \textbf{45.33} & \underline{64.26} & \textbf{66.37} & \underline{78.70} & \textbf{24.30} & \textbf{49.83} \\
\bottomrule
\end{tabular}
}
\caption{Class-wise comparison on UAVid. Overall is computed over all six foreground classes; seen and unseen scores are averaged over their corresponding category groups.}
\label{tab:uavid_classwise}
\vspace{-0.5em}
\end{table*}

\cref{tab:uavid_classwise} evaluates seen and unseen category generalization on UAVid. Building, Road, and Tree are treated as seen classes, while Low vegetation, Car, and Human are treated as unseen. Pi-Seg achieves the best seen mIoU and the strongest unseen performance in both mIoU and mACC. The gains therefore extend beyond fitting seen categories and improve transfer to unseen semantics.

The improvement on unseen classes is important because open-vocabulary segmentation is not merely a closed-set segmentation problem with text labels. A method may perform well on categories that are visually or semantically close to the training distribution, but still fail when queried with less frequent or visually diverse categories. The UAVid results show that Pi-Seg improves both the seen group and the unseen group, suggesting that semantic perturbation helps preserve transferability rather than only improving source-category fitting.

\begin{table*}[tb]
\centering
\small
\resizebox{0.95\linewidth}{!}{
\begin{tabular}{lcccccccccccccc}
\toprule
\multirow{2}{*}{Method}
& \multicolumn{2}{c}{Vegetation}
& \multicolumn{2}{c}{Building}
& \multicolumn{2}{c}{Road}
& \multicolumn{2}{c}{Vehicle}
& \multicolumn{2}{c}{Overall}
& \multicolumn{2}{c}{Seen}
& \multicolumn{2}{c}{Unseen} \\
\cmidrule(lr){2-3} \cmidrule(lr){4-5} \cmidrule(lr){6-7}
\cmidrule(lr){8-9} \cmidrule(lr){10-11} \cmidrule(lr){12-13} \cmidrule(lr){14-15}
& IoU & ACC & IoU & ACC & IoU & ACC & IoU & ACC & mIoU & mACC & mIoU & mACC & mIoU & mACC \\
\midrule
CAT-Seg  & \underline{64.17} & \underline{76.35} & \underline{73.32} & \textbf{97.68} & \underline{47.66} & 78.78 & \underline{40.16} & \underline{72.34} & \underline{56.33} & \underline{81.29} & \underline{53.71} & \underline{82.93} & \underline{64.17} & \underline{76.35} \\
RSKT-Seg & 33.62 & 36.40 & 71.66 & 91.60 & 27.17 & \underline{85.88} & 5.47 & 5.57 & 34.48 & 54.86 & 34.77 & 61.02 & 33.62 & 36.40 \\
Pi-Seg   & \textbf{78.26} & \textbf{92.73} & \textbf{86.19} & \underline{93.79} & \textbf{62.58} & \textbf{89.07} & \textbf{46.76} & \textbf{77.58} & \textbf{68.45} & \textbf{88.29} & \textbf{65.18} & \textbf{86.81} & \textbf{78.26} & \textbf{92.73} \\
\bottomrule
\end{tabular}
}
\caption{Class-wise comparison on UDD5 after removing the Background/Clutter category. Vegetation is treated as unseen, while Building, Road, and Vehicle are treated as seen.}
\label{tab:udd5_classwise_seen_unseen}
\vspace{-0.5em}
\end{table*}

The additional UDD5 analysis in \cref{tab:udd5_classwise_seen_unseen} provides a second seen/unseen evaluation setting. Vegetation is treated as the unseen category, while Building, Road, and Vehicle are treated as seen categories. Pi-Seg substantially improves both the overall score and the unseen Vegetation category, further supporting the effectiveness of semantic perturbation for open-vocabulary transfer.

\subsection{Heavy-Tailed Perturbation}
\label{sec:suppl_heavy_tailed}

\begin{table}[tb]
\centering
\small
\resizebox{0.95\linewidth}{!}{
\begin{tabular}{ccccc}
\toprule
Image noise & Text noise & Student-$t$ df ($\nu$) & m-mIoU & m-mACC \\
\midrule
Gaussian    & Student-$t$ & 3   & 38.86 & 55.38 \\
Gaussian    & Student-$t$ & 5   & 38.50 & 55.71 \\
Gaussian    & Student-$t$ & 10  & \textbf{40.12} & 56.91 \\
\midrule
Student-$t$ & Gaussian    & 3  & 39.32 & 56.96 \\
Student-$t$ & Gaussian    & 5  & 38.82 & \textbf{57.60} \\
Student-$t$ & Gaussian    & 10 & \textbf{39.48} & 57.42 \\
\bottomrule
\end{tabular}
}
\caption{Effect of the Student-$t$ degrees of freedom.}
\label{tab:student_t_df_tpami}
\vspace{-0.5em}
\end{table}

\cref{tab:student_t_df_tpami} studies Student-$t$ perturbations with different degrees of freedom $\nu$. Moderate heavy-tailed perturbations generally outperform more aggressive settings. Excessive perturbation may disrupt feature alignment and optimization, supporting the hypothesis that effective perturbation should expand the transferable feature neighborhood in a controlled manner.

Student-$t$ noise provides a useful way to test whether occasional larger perturbations are beneficial. Smaller degrees of freedom produce heavier tails and therefore more extreme samples, while larger degrees of freedom behave closer to Gaussian noise. The results suggest that the best setting is not the most aggressive one. Instead, a moderate heavy-tailed distribution can improve exploration of the feature space while avoiding unstable perturbations that damage semantic alignment.

\subsection{Robustness Across Image Resolutions}
\label{sec:suppl_resolution_robustness}

\begin{table}[tb]
\centering
\footnotesize

\begin{subtable}[t]{\linewidth}
\centering
\setlength{\tabcolsep}{8pt}
\resizebox{\linewidth}{!}{
\begin{tabular}{lll}
\toprule
Dataset & Resolution (Width $\times$ Height) & Group \\
\midrule
DLRSD        & 256 $\times$ 256         & Low-resolution \\
FLAIR        & 512 $\times$ 512         & Low-resolution \\
iSAID        & 256 $\times$ 256         & Low-resolution \\
LoveDA       & 1024 $\times$ 1024       & High-resolution \\
OpenEarthMap & 981.94 $\times$ 981.34   & High-resolution \\
Potsdam      & 256 $\times$ 256         & Low-resolution \\
UAVid        & 3934.81 $\times$ 2160.00 & High-resolution \\
UDD5         & 4003.20 $\times$ 2748.00 & High-resolution \\
Vaihingen    & 256 $\times$ 256         & Low-resolution \\
VDD          & 4000 $\times$ 3000       & High-resolution \\
\bottomrule
\end{tabular}
}
\caption{Dataset resolution grouping.}
\label{tab:dataset_resolution_grouping}
\end{subtable}

\vspace{0.8em}

\begin{subtable}[t]{\linewidth}
\centering
\setlength{\tabcolsep}{6pt}
\resizebox{\linewidth}{!}{
\begin{tabular}{lcccc}
\toprule
\multirow{2}{*}{Method} & \multicolumn{2}{c}{High-Resolution} & \multicolumn{2}{c}{Low-Resolution} \\
\cmidrule(lr){2-3} \cmidrule(lr){4-5}
 & m-mIoU & m-mACC & m-mIoU & m-mACC \\
\midrule
SegEarth-OV\cite{li2025segearth}$_{\textcolor{gray}{\text{\textit{CVPR2025}}}}$
& 30.04 & 44.45 & 21.35 & 37.66 \\
SCAN\cite{liu2024open}$_{\textcolor{gray}{\text{\textit{CVPR2024}}}}$
& 29.10 & 44.90 & 19.50 & 36.17 \\
SAN\cite{xu2023san}$_{\textcolor{gray}{\text{\textit{CVPR2023}}}}$
& 31.30 & 47.18 & 20.99 & 37.09 \\
FCCLIP\cite{yu2023convolutions}$_{\textcolor{gray}{\text{\textit{NeurIPS2023}}}}$
& 37.61 & 58.62 & 22.26 & 38.78 \\
CAT-SEG\cite{cho2024cat}$_{\textcolor{gray}{\text{\textit{CVPR2024}}}}$
& \underline{42.41} & \underline{61.71} & 33.50 & 49.91 \\
OVRS\cite{cao2025open}$_{\textcolor{gray}{\text{\textit{TGRS2025}}}}$
& 42.07 & 59.67 & 32.84 & 49.34 \\
RSKT-SEG\cite{li2026exploring}$_{\textcolor{gray}{\text{\textit{AAAI2026}}}}$
& 37.40 & 54.50 & \underline{33.85} & \textbf{51.68} \\
Pi-Seg
& \textbf{44.50} & \textbf{62.93} & \textbf{34.46} & \underline{50.87} \\
\bottomrule
\end{tabular}
}
\caption{Performance under different pixel regimes.}
\label{tab:high_low_resolution_base}
\end{subtable}
\caption{Resolution grouping and performance comparison under different pixel regimes.}
\label{tab:resolution_analysis_ab}
\vspace{-0.5em}
\end{table}

We group the evaluation datasets into low- and high-resolution subsets according to their native image sizes, as summarized in \cref{tab:resolution_analysis_ab}(a). The averaged results in \cref{tab:resolution_analysis_ab}(b) show that Pi-Seg achieves the best high-resolution m-mIoU and m-mACC, as well as the best low-resolution m-mIoU. These results demonstrate robust generalization across image scales, with particularly strong performance on high-resolution remote-sensing imagery.

High-resolution remote sensing images often contain fine object boundaries, small vehicles, narrow roads, and dense building structures. Low-resolution images, in contrast, may lose fine details but require stronger semantic abstraction. The consistent results across both groups indicate that Pi-Seg does not rely on a single pixel regime. The semantic perturbation mechanism improves dense alignment in both detailed high-resolution scenes and more compact low-resolution scenes.

\subsection{Additional Qualitative Comparison}
\label{sec:suppl_additional_qualitative}

\begin{figure*}[tb]
    \centering
    \includegraphics[width=\linewidth]{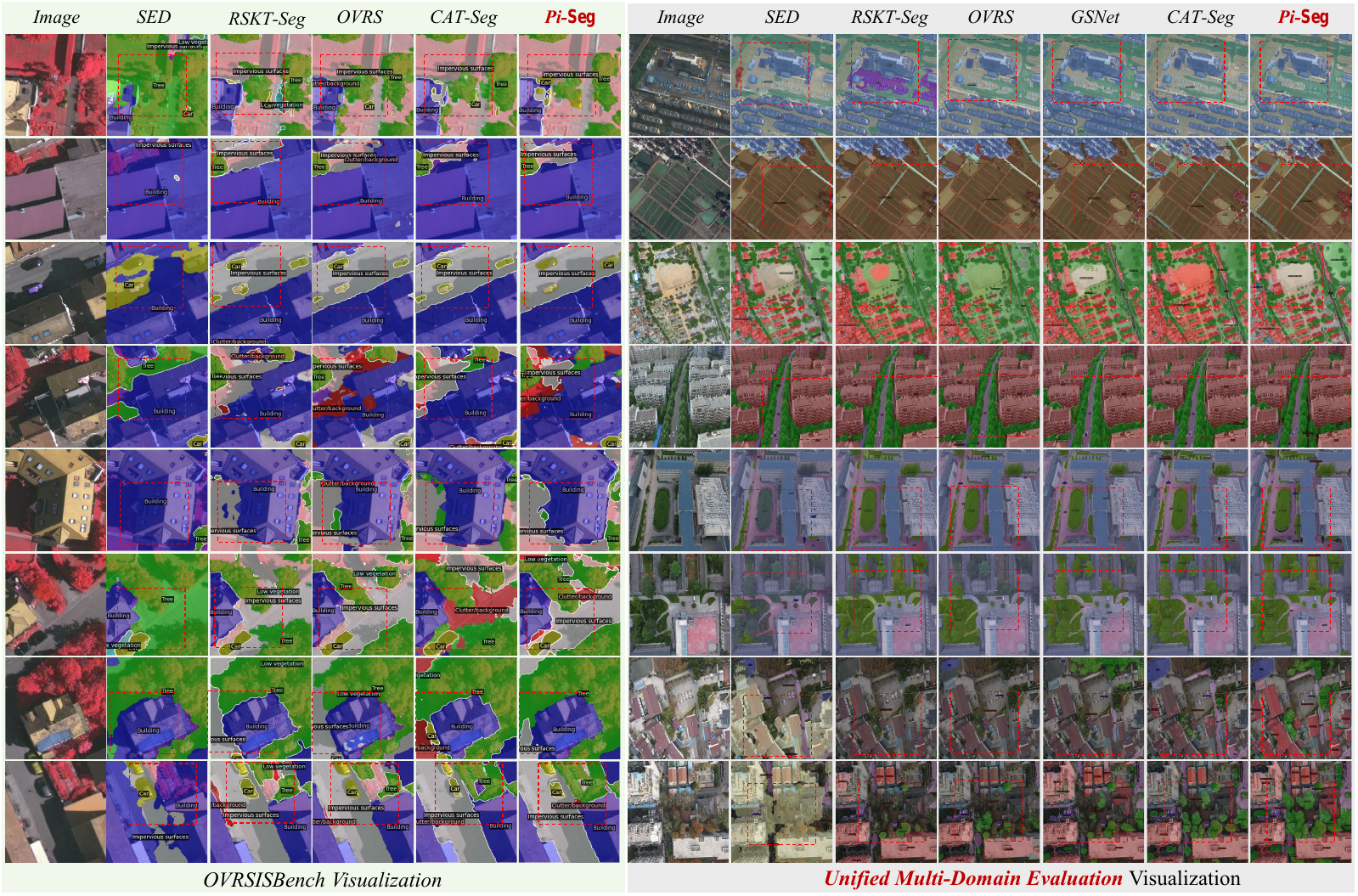}
    \caption{\textbf{Additional qualitative comparison.} Pi-Seg produces more complete regions and fewer background activations across representative OVRSIS examples.}
    \label{fig:suppl_vis_segmap_bench}
    \vspace{-0.5em}
\end{figure*}

\cref{fig:suppl_vis_segmap_bench} provides additional qualitative comparisons across representative OVRSIS examples. The visual results show that Pi-Seg often produces more spatially complete regions, especially for large objects with complex boundaries. Competing methods may produce fragmented masks or activate background regions with similar textures. In contrast, Pi-Seg tends to preserve the target extent while suppressing unrelated regions. This behavior is consistent with the quantitative gains reported in the main benchmark tables.

The improvement is particularly visible in scenes where target objects occupy irregular shapes or are surrounded by visually similar background. Such cases are common in remote-sensing imagery because many categories share materials, textures, or contextual layouts. By perturbing both textual and visual representations during training, Pi-Seg learns a more tolerant yet semantically constrained matching space, which helps reduce brittle pixel--text alignment at inference time.

\subsection{Definition of the $\Delta$-Correlation Statistics}
\label{sec:suppl_delta_corr_stats}

We use four $\Delta$-correlation statistics to quantify how the learned visual perturbation changes dense pixel--text correspondence during training. Let $V^{\mathrm{pre}}_{b,u}$ and $V^{\mathrm{post}}_{b,u}$ denote the visual feature at spatial location $u$ before and after Image-SPM, respectively. Let $T'_{b,c,p}$ denote the text embedding of category $c$ under prompt $p$ that is used by the model at the current training step. When Text-SPM is enabled, $T'$ already includes the learned text perturbation. The same $T'$ is used for the pre- and post-Image-SPM correlations; consequently, the following difference isolates the incremental effect of Image-SPM under the current text conditioning.

For image $b$, prompt $p$, category $c$, and feature-grid location $u$, we compute
\begin{align}
C^{\mathrm{pre}}_{b,p,c,u}
&=
\left\langle
\frac{V^{\mathrm{pre}}_{b,u}}{\|V^{\mathrm{pre}}_{b,u}\|_2},
\frac{T'_{b,c,p}}{\|T'_{b,c,p}\|_2}
\right\rangle, \\
C^{\mathrm{post}}_{b,p,c,u}
&=
\left\langle
\frac{V^{\mathrm{post}}_{b,u}}{\|V^{\mathrm{post}}_{b,u}\|_2},
\frac{T'_{b,c,p}}{\|T'_{b,c,p}\|_2}
\right\rangle,
\end{align}
and define the perturbation-induced correlation change as
\begin{equation}
\Delta C_{b,p,c,u}
=
C^{\mathrm{post}}_{b,p,c,u}
-
C^{\mathrm{pre}}_{b,p,c,u}.
\label{eq:suppl_delta_corr}
\end{equation}
The implementation first averages $\Delta C$ over the $P$ prompt templates,
\begin{equation}
\overline{\Delta C}_{b,c,u}
=
\frac{1}{P}
\sum_{p=1}^{P}
\Delta C_{b,p,c,u},
\label{eq:suppl_prompt_avg_delta_corr}
\end{equation}
and bilinearly resizes the resulting category-wise maps from the feature grid to the ground-truth mask resolution. We denote the resized value at pixel $i$ by $\widetilde{\Delta C}_{b,c,i}$.

Let $Y_{b,i}$ be the ground-truth category at pixel $i$, let $y_{\mathrm{ign}}$ be the ignored label, and define the valid-pixel set and the categories present in image $b$ as
\begin{equation}
\Omega_b
=
\{i:Y_{b,i}\neq y_{\mathrm{ign}}\},
\qquad
\mathcal{C}_b
=
\{Y_{b,i}:i\in\Omega_b\}.
\end{equation}
For every category $c\in\mathcal{C}_b$, pixels belonging to $c$ form the ground-truth region, whereas all other valid pixels form its non-ground-truth region. The two elementary statistics are therefore
\begin{align}
\mathrm{gt\_in\_mean}
&=
\frac{
\displaystyle
\sum_b\sum_{c\in\mathcal{C}_b}
\sum_{i\in\Omega_b}
\mathbf{1}[Y_{b,i}=c]\,
\widetilde{\Delta C}_{b,c,i}
}{
\displaystyle
\sum_b\sum_{c\in\mathcal{C}_b}
\sum_{i\in\Omega_b}
\mathbf{1}[Y_{b,i}=c]
},
\label{eq:suppl_gt_in_mean}\\
\mathrm{non\_gt\_mean}
&=
\frac{
\displaystyle
\sum_b\sum_{c\in\mathcal{C}_b}
\sum_{i\in\Omega_b}
\mathbf{1}[Y_{b,i}\neq c]\,
\widetilde{\Delta C}_{b,c,i}
}{
\displaystyle
\sum_b\sum_{c\in\mathcal{C}_b}
\sum_{i\in\Omega_b}
\mathbf{1}[Y_{b,i}\neq c]
}.
\label{eq:suppl_non_gt_mean}
\end{align}
These are element-weighted means over all valid images, present categories, and pixels. In distributed training, the corresponding sums and counts are aggregated across all GPUs before the global means are computed. Images without valid pixels are skipped.

A positive $\mathrm{gt\_in\_mean}$ indicates that Image-SPM increases the correlation between a category and pixels inside its ground-truth region. A negative $\mathrm{non\_gt\_mean}$ indicates that the same category response is suppressed outside its ground-truth region. We further report
\begin{align}
\mathrm{gap}
&=
\mathrm{gt\_in\_mean}
-
\mathrm{non\_gt\_mean},
\label{eq:suppl_delta_gap}\\
\mathrm{align\_ratio}
&=
\frac{\mathrm{gt\_in\_mean}}
{|\mathrm{non\_gt\_mean}|+\varepsilon},
\qquad
\varepsilon=10^{-6}.
\label{eq:suppl_align_ratio}
\end{align}
The gap measures the separation between target-region enhancement and off-target change. The alignment ratio measures how strongly the target-region increase dominates the magnitude of the average off-target change. A value of one is a balance reference rather than the no-perturbation value, and this ratio is interpreted jointly with the signs of $\mathrm{gt\_in\_mean}$ and $\mathrm{non\_gt\_mean}$ and with the gap.

The statistics are logged once per optimization step. For the curves in Figure~6, we display the raw step-wise measurements together with trailing rolling means. A window of 500 steps is used for $\mathrm{gt\_in\_mean}$, $\mathrm{non\_gt\_mean}$, and $\mathrm{gap}$; a window of 800 steps is used for $\mathrm{align\_ratio}$. To reduce the visual influence of rare denominator-induced spikes, the raw alignment-ratio values are clipped to their 5th--95th percentile range before smoothing. This processing is used only for visualization and does not affect model training or the logged raw statistics.

\section{More Failure Cases}
\label{sec:suppl_failure_cases}

\begin{figure}[tb]
    \centering
    \includegraphics[width=0.95\linewidth]{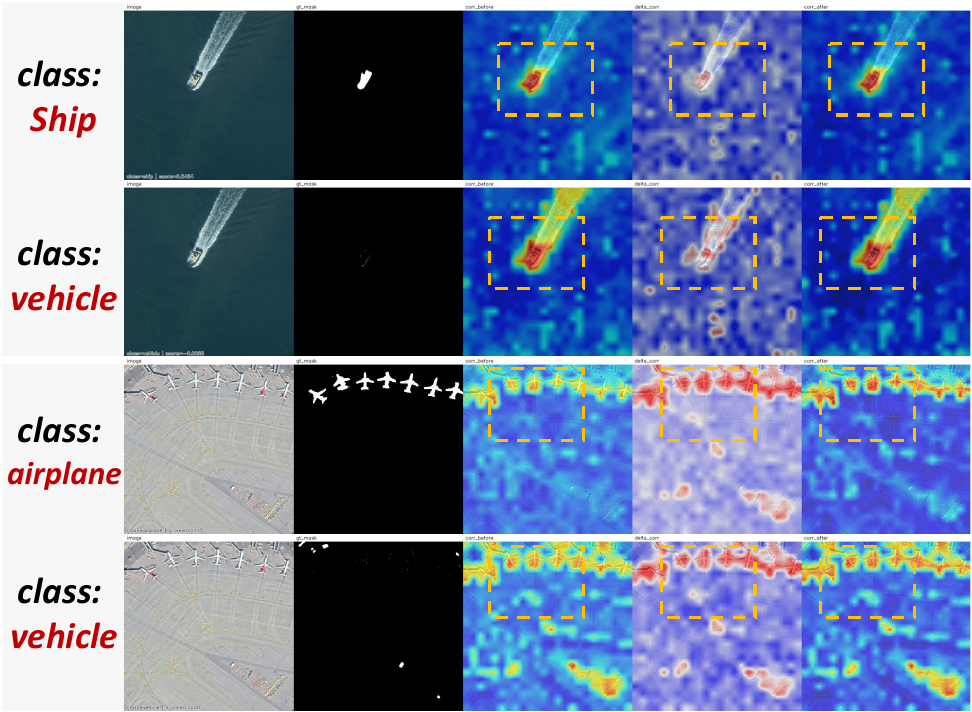}
    \caption{\textbf{Representative failure cases.} Pi-Seg may confuse fine-grained categories with broader semantic concepts under ambiguous visual evidence.}
    \label{fig:failure_cases_tpami1}
    \vspace{-0.5em}
\end{figure}

\cref{fig:failure_cases_tpami1} presents representative failure cases. Pi-Seg may confuse fine-grained categories such as \emph{ship} and \emph{airplane} with the broader category \emph{vehicle}. This limitation mainly stems from coarse semantic prototypes derived from short class prompts, which provide insufficient distinctions among visually related concepts.

Although semantic perturbation improves robustness and transferability, it does not introduce fine-grained semantic supervision or high-level contextual reasoning by itself. Future work may incorporate richer textual descriptions, hierarchical category definitions, and stronger context modeling to improve fine-grained category discrimination.

Another limitation is that the proposed perturbation mechanism operates in the feature space and depends on the quality of the pretrained vision--language representation. If a category is poorly represented by the original language embedding or visually rare in the training data, perturbation alone may not fully recover the missing semantic knowledge. This suggests that future OVRSIS research may benefit from combining feature-space regularization with richer text prompts, external geographic knowledge, or multi-modal remote sensing signals.

\section{Variational Interpretation of Pi-Seg}
\label{sec:suppl_variational_piseg}

This section provides a variational interpretation of why Pi-Seg can improve dense open-vocabulary segmentation. The analysis follows the idea of Variational Positive-Incentive Noise, where learnable stochastic perturbations are optimized to benefit the model rather than simply corrupting its features. Here, we adapt this view to the pixel--text cost aggregation formulation used in OVRSIS.

\subsection{Dense Objective with Perturbed Features}
\label{sec:suppl_dense_objective}

Let $\mathcal{P}$ be the set of valid pixel locations and $y_i \in \{1,\ldots,N\}$ be the ground-truth label of pixel $i$. Given dense visual features $V=\{V_i\}_{i\in\mathcal{P}}$ and text embeddings $T=\{T_n\}_{n=1}^{N}$, a cost-based OVRSIS model first computes the pixel--text cost map:
\begin{equation}
\label{eq:suppl_clean_cost}
C_{i,n}
=
\frac{V_i^\top T_n}
{\|V_i\|_2\|T_n\|_2}.
\end{equation}
The cost map is then processed by a spatial/class-wise aggregation module and an upsampling decoder:
\begin{equation}
\label{eq:suppl_clean_logit}
Z
=
\mathcal{D}_{\theta}
\big(
    \mathcal{F}_{\mathrm{agg}}(C)
\big),
\end{equation}
where $\mathcal{F}_{\mathrm{agg}}$ denotes cost aggregation and $\mathcal{D}_{\theta}$ denotes the decoder. The pixel-wise segmentation loss is
\begin{equation}
\label{eq:suppl_seg_loss}
\mathcal{L}_{\mathrm{seg}}(Z,Y)
=
\frac{1}{|\mathcal{P}|}
\sum_{i\in\mathcal{P}}
-\log
\frac{\exp(Z_{i,y_i})}
{\sum_{n=1}^{N}\exp(Z_{i,n})}.
\end{equation}

Pi-Seg introduces stochastic perturbations before cost construction:
\begin{equation}
\label{eq:suppl_feature_perturb}
\hat{T}_n = T_n+\epsilon^t_n,
\qquad
\hat{V}_i = V_i+\epsilon^v_i ,
\end{equation}
where $\epsilon^t_n$ is generated by Text-SPM and $\epsilon^v_i$ is generated by Image-SPM. The perturbed cost map and logits are
\begin{align}
\label{eq:suppl_perturbed_cost}
\hat{C}_{i,n}
&=
\frac{\hat{V}_i^\top \hat{T}_n}
{\|\hat{V}_i\|_2\|\hat{T}_n\|_2},
\\
\label{eq:suppl_perturbed_logit}
\hat{Z}
&=
\mathcal{D}_{\theta}
\big(
    \mathcal{F}_{\mathrm{agg}}(\hat{C})
\big).
\end{align}
Therefore, Pi-Seg does not inject perturbations into raw images. Instead, it perturbs the vision--language feature space where dense pixel--text matching is performed.

\subsection{An Idealized Variational Perspective}
\label{sec:suppl_vpin_objective}

Let $\epsilon=(\epsilon^v,\epsilon^t)$ denote the joint perturbation, and let $p_0(\epsilon)$ be a simple prior distribution, such as a standard Gaussian. Following the variational positive-incentive perspective~\cite{zhang2025variational}, an ideal perturbation mechanism should assign greater probability to feature variations that yield a low segmentation loss. We introduce the following soft positive-incentive objective for interpretation:
\begin{equation}
\label{eq:suppl_positive_objective}
\mathcal{J}^{+}
=
-\log
\mathbb{E}_{\epsilon\sim p_0}
\left[
\exp
\big(
-\mathcal{L}_{\mathrm{seg}}(\hat{Z},Y)
\big)
\right].
\end{equation}
This quantity becomes small when the prior perturbation space contains samples that produce a low segmentation loss. Directly evaluating it is generally intractable because it requires integration over the full perturbation space.

To relate this idealized objective to a learnable proposal, consider the data-dependent perturbation distribution $q_{\phi,\psi}(\epsilon\mid V,T)$ parameterized by Text-SPM and Image-SPM. Inserting $q_{\phi,\psi}$ into \cref{eq:suppl_positive_objective} gives
\begin{align}
\mathcal{J}^{+}
&=
-\log
\int
q_{\phi,\psi}(\epsilon\mid V,T)
\frac{p_0(\epsilon)}
     {q_{\phi,\psi}(\epsilon\mid V,T)}
\exp
\big(
-\mathcal{L}_{\mathrm{seg}}(\hat{Z},Y)
\big)
d\epsilon .
\label{eq:suppl_insert_q}
\end{align}
Using Jensen's inequality, we obtain
\begin{equation}
\label{eq:suppl_variational_bound}
\mathcal{J}^{+}
\leq
\mathbb{E}_{q_{\phi,\psi}}
\left[
\mathcal{L}_{\mathrm{seg}}(\hat{Z},Y)
\right]
+
D_{\mathrm{KL}}
\left(
q_{\phi,\psi}(\epsilon\mid V,T)
\|p_0(\epsilon)
\right).
\end{equation}
Equation~\eqref{eq:suppl_variational_bound} is an idealized prior-regularized upper bound, rather than the objective optimized in our implementation. It connects Pi-Seg to variational positive-incentive noise by interpreting $q_{\phi,\psi}$ as a data-dependent proposal over task-favorable perturbations, but Pi-Seg does not explicitly minimize the KL divergence or enforce posterior matching to $p_0$.

The oracle Gibbs distribution associated with the idealized objective is
\begin{equation}
\label{eq:suppl_optimal_noise}
q^{*}(\epsilon\mid V,T,Y)
\propto
p_0(\epsilon)
\exp
\big(
-\mathcal{L}_{\mathrm{seg}}(\hat{Z},Y)
\big).
\end{equation}
This label-dependent distribution serves only as an oracle reference: it assigns greater mass to perturbations that reduce the dense segmentation loss, but it is neither instantiated nor explicitly matched by Pi-Seg.

In practice, Pi-Seg directly optimizes the empirical objective
\begin{equation}
\label{eq:suppl_practical_objective}
\min_{\theta,\phi,\psi}
\mathbb{E}_{(I,Y)}
\mathbb{E}_{\epsilon\sim q_{\phi,\psi}}
\left[
\mathcal{L}_{\mathrm{seg}}
\left(
\mathcal{D}_{\theta}
\big(
\mathcal{F}_{\mathrm{agg}}(\hat{C})
\big),
Y
\right)
\right].
\end{equation}
No explicit KL term is used in \cref{eq:suppl_practical_objective}. Therefore, the variational formulation above should be understood as an interpretation rather than an exact derivation of the implemented training objective. The perturbation scale is controlled through initialization and scale parameterization, and its practical stability is evaluated through the sensitivity and random-seed analyses in \cref{sec:suppl_hyperparameter_sensitivity,sec:suppl_seed_stability}. Minimizing \cref{eq:suppl_practical_objective} discourages perturbations that increase the supervised loss on average, but it does not explicitly enforce prior matching.

\subsection{Effect on Pixel--Text Cost Aggregation}
\label{sec:suppl_pixel_text_cost_effect}

We next show how feature perturbation changes the dense cost map. Let
$\bar{V}_i=V_i/\|V_i\|_2$ and $\bar{T}_n=T_n/\|T_n\|_2$. For small perturbations, the perturbed cosine similarity can be approximated by first-order expansion:
\begin{align}
\hat{C}_{i,n}
\approx
C_{i,n}
&+
\left\langle
A^v_{i,n},
\epsilon^v_i
\right\rangle
+
\left\langle
A^t_{i,n},
\epsilon^t_n
\right\rangle ,
\label{eq:suppl_cost_taylor_short}
\end{align}
where
\begin{align}
A^v_{i,n}
&=
\frac{(I-\bar{V}_i\bar{V}_i^\top)\bar{T}_n}
{\|V_i\|_2},
&
A^t_{i,n}
&=
\frac{(I-\bar{T}_n\bar{T}_n^\top)\bar{V}_i}
{\|T_n\|_2}.
\label{eq:suppl_cost_taylor_terms}
\end{align}
This approximation shows that Image-SPM changes the cost response at each spatial location, while Text-SPM changes the category prototype used by all pixels. Therefore, Pi-Seg jointly expands the visual and textual neighborhoods involved in dense matching.

Cost aggregation further propagates these perturbation-induced changes. Let
$\mathcal{G}_{\theta}=\mathcal{D}_{\theta}\circ\mathcal{F}_{\mathrm{agg}}$.
For a small cost variation $\delta C$, the logit variation can be approximated as
\begin{equation}
\label{eq:suppl_aggregation_propagation}
\delta Z_{i,n}
\approx
\sum_{k\in\mathcal{P}}
\sum_{m=1}^{N}
\frac{\partial Z_{i,n}}
     {\partial C_{k,m}}
\delta C_{k,m}.
\end{equation}
Thus, a perturbation at one pixel or category channel can affect neighboring regions and competing categories through the aggregation module. This is important for fine-grained segmentation, where correct prediction depends on both local evidence and contextual consistency.

\subsection{Positive Incentive as Margin Improvement}
\label{sec:suppl_positive_margin}

For pixel $i$, define the target-vs-distractor margin as
\begin{equation}
\label{eq:suppl_margin}
M_i(Z)
=
Z_{i,y_i}
-
\log
\sum_{n\neq y_i}
\exp(Z_{i,n}).
\end{equation}
The pixel-wise cross-entropy can be rewritten as
\begin{equation}
\label{eq:suppl_margin_loss}
\ell_i(Z,Y)
=
\log
\big(
1+\exp(-M_i(Z))
\big).
\end{equation}
Thus, a larger target-vs-distractor margin corresponds to a smaller pixel-wise loss. We use
\begin{equation}
\label{eq:suppl_positive_margin}
\mathbb{E}_{q_{\phi,\psi}}
\left[
M_i(\hat{Z})-M_i(Z)
\right]
>0
\end{equation}
as a diagnostic condition for positive-incentive behavior at pixel $i$. This condition compares the expected perturbed margin with the clean margin; it is not a guaranteed consequence of the empirical objective in \cref{eq:suppl_practical_objective}, which directly optimizes only the perturbed segmentation loss.

Let $\mu^v_i=\mathbb{E}_{q}[\epsilon^v_i]$ and $\mu^t_n=\mathbb{E}_{q}[\epsilon^t_n]$. A local first-order expansion gives
\begin{equation}
\label{eq:suppl_loss_change}
\mathbb{E}_{q}[\delta \ell_i]
\approx
\left\langle
\nabla_{V_i}\ell_i,
\mu^v_i
\right\rangle
+
\sum_{n=1}^{N}
\left\langle
\nabla_{T_n}\ell_i,
\mu^t_n
\right\rangle .
\end{equation}
Under reparameterized sampling, gradients from \cref{eq:suppl_practical_objective} propagate through the perturbation means predicted by the two generators. Locally, minimizing \cref{eq:suppl_loss_change} encourages these means to align with loss-reducing directions. Because the means are outputs of Text-SPM and Image-SPM rather than independent per-pixel optimization variables, this expression provides an optimization intuition rather than an explicit update rule.

The variance terms control how broadly the neighborhood is explored. Collecting the visual and textual perturbations relevant to pixel $i$ into a joint variable with mean $\mu_i$ and covariance $\Sigma_i$, a second-order approximation gives
\begin{equation}
\label{eq:suppl_second_order}
\mathbb{E}_{q}[\ell_i(\hat{Z})]
\approx
\ell_i(Z)
+
g_i^\top\mu_i
+
\frac{1}{2}
\mathrm{Tr}
\left[
H_i
\left(
\Sigma_i+\mu_i\mu_i^\top
\right)
\right],
\end{equation}
where $g_i$ and $H_i$ are the gradient and Hessian with respect to the joint perturbation. This approximation explains why controlled perturbation is important: moderate variance explores nearby directions, whereas excessive variance may cross semantic boundaries and damage pixel--text alignment.

\subsection{Implication for Pi-Seg}
\label{sec:suppl_piseg_implication}

The analysis above provides a local interpretation of Pi-Seg rather than a formal guarantee. Text-SPM expands fixed textual prototypes into task-optimized semantic neighborhoods, while Image-SPM generates text-conditioned, spatially varying visual perturbations. Cost aggregation propagates the induced changes across spatial locations and category channels. Pixel-wise supervision directly favors low-loss perturbed samples, and the relative margin criterion in \cref{eq:suppl_positive_margin} is evaluated empirically through the training dynamics reported in the main paper. Together, these observations explain why the learned perturbations differ from arbitrary random noise and can improve dense pixel--text alignment under category and domain shifts.

\section{Open-Vocabulary Generalization Interpretation}
\label{sec:open_vocabulary_generalization}

The open-vocabulary capability of Pi-Seg originates from the shared
vision--language space of the frozen pretrained encoders. For a category
name $c$, the text encoder produces
\begin{equation}
    T_c
    =
    E_{\mathrm{text}}
    \bigl(
        \operatorname{prompt}(c)
    \bigr),
    \qquad
    c\in
    \mathcal{C}_{\mathrm{s}}
    \cup
    \mathcal{C}_{\mathrm{u}},
    \label{eq:ov_text_prototype}
\end{equation}
where $\mathcal{C}_{\mathrm{s}}$ and $\mathcal{C}_{\mathrm{u}}$ denote
the seen and unseen category sets, respectively, with
$\mathcal{C}_{\mathrm{s}}\cap\mathcal{C}_{\mathrm{u}}=\varnothing$.
This shared embedding space provides an open category interface, but it
does not guarantee that downstream segmentation training will preserve
reliable correspondence for unseen categories. Pi-Seg does not create
unseen-category knowledge; instead, it regularizes the downstream
matching function so that the open semantic geometry inherited from the
pretrained model is less prone to overfitting to seen categories.

\subsection{From Prototype Points to Semantic Neighborhoods}
\label{sec:ov_semantic_support}

Conventional deterministic matching receives supervision only at the
finite set of seen textual prototypes
\begin{equation}
    \mathcal{S}_{\mathrm{point}}
    =
    \left\{
        T_c
        \mid
        c\in\mathcal{C}_{\mathrm{s}}
    \right\}.
    \label{eq:ov_point_support}
\end{equation}
Although these prototypes lie in an open semantic space, the downstream
cost aggregation and decoder may adapt to their exact coordinates,
prompt forms, and source-domain statistics. This can produce strong
responses at seen prototypes but unstable behavior at nearby textual
coordinates, resulting in seen-prototype overfitting.

Pi-Seg perturbs the visual and textual features during training:
\begin{equation}
    \widehat{V}_i
    =
    V_i+\epsilon_i^{v},
    \qquad
    \widehat{T}_c
    =
    T_c+\epsilon_c^{t},
    \qquad
    \epsilon
    \sim
    q_{\phi,\psi}(\epsilon\mid V,T).
    \label{eq:ov_perturbed_features}
\end{equation}
For a confidence level $\rho\in(0,1)$, let
$\mathcal{E}_c(\rho)$ be a bounded high-probability region of the marginal
text-perturbation distribution satisfying
\begin{equation}
    \Pr_{\epsilon_c^t\sim q_{\phi,\psi}^{t}(\cdot\mid V,T)}
    \left[
        \epsilon_c^t\in\mathcal{E}_c(\rho)
    \right]
    \geq \rho .
    \label{eq:ov_high_probability_region}
\end{equation}
We define the semantic neighborhood covered around category $c$ as
\begin{equation}
\begin{aligned}
    \mathcal{U}_c(\rho)
    &=
    \{T_c\}
    \cup
    \left\{
        T_c+\epsilon_c^t
        :
        \epsilon_c^t\in\mathcal{E}_c(\rho)
    \right\},
    \\
    \mathcal{S}_{\rho}
    &=
    \bigcup_{c\in\mathcal{C}_{\mathrm{s}}}
    \mathcal{U}_c(\rho).
\end{aligned}
\label{eq:ov_distributional_support}
\end{equation}
The bounded high-probability region is used instead of the full mathematical
support of $q_{\phi,\psi}^{t}$, because Gaussian and Student-$t$
distributions have unbounded support and should not be interpreted as
covering the entire text-embedding space.

The practical objective
\begin{equation}
\begin{aligned}
    \min_{\theta,\phi,\psi}
    \quad
    \mathbb{E}_{(I,Y)}
    \mathbb{E}_{\epsilon\sim q_{\phi,\psi}}
    \left[
        \mathcal{L}_{\mathrm{seg}}
        \left(
            A_{\theta}
            \bigl(
                V+\epsilon^{v},
                T+\epsilon^{t}
            \bigr),
            Y
        \right)
    \right]
    \label{eq:ov_distributional_objective}
\end{aligned}
\end{equation}
samples local visual--textual variations rather than supervising the
alignment operator $A_{\theta}(V,T)$ only at isolated seen prototypes.
Text-SPM broadens the semantic region explored during training, whereas
Image-SPM exposes the model to spatially varying appearance changes. Their
combination discourages memorization of fixed category coordinates and
promotes a smoother matching rule that may be reused when the inference
vocabulary changes.

\subsection{Support-Distance View of Unseen Transfer}
\label{sec:ov_support_distance}

Let $A^{*}(V,T)$ denote an ideal correspondence function and
$A_{\theta}(V,T)$ the learned one. Assume that both functions are locally
Lipschitz continuous with respect to $T$, with constants $L_{*}$ and
$L_{\theta}$, and that the alignment error on the bounded covered region
$\mathcal{S}_{\rho}$ is at most $\varepsilon_{\mathrm{train}}$. For an
unseen prototype $T_u$, applying the triangle inequality through the nearest
point in $\mathcal{S}_{\rho}$ gives
\begin{equation}
\begin{aligned}
    \left\|
        A_{\theta}(V,T_u)
        -
        A^{*}(V,T_u)
    \right\|
    \leq
    \varepsilon_{\mathrm{train}}
    +
    (L_{\theta}+L_{*})
    \operatorname{dist}
    \left(
        T_u,
        \mathcal{S}_{\rho}
    \right).
    \label{eq:ov_unseen_bound}
\end{aligned}
\end{equation}
Because each neighborhood explicitly contains its original prototype,
$\mathcal{S}_{\mathrm{point}}\subseteq\mathcal{S}_{\rho}$, and therefore
\begin{equation}
    \operatorname{dist}
    \left(
        T_u,
        \mathcal{S}_{\rho}
    \right)
    \leq
    \operatorname{dist}
    \left(
        T_u,
        \mathcal{S}_{\mathrm{point}}
    \right).
    \label{eq:ov_distance_reduction}
\end{equation}
Under the stated local assumptions, this provides an interpretation of the
open-vocabulary benefit: distributional training can expand the semantic
region covered by pixel-level supervision and reduce the distance between
an unseen prototype and the covered training region. When $T_u$ lies near
such a region, it may reuse the class-agnostic alignment behavior learned
there. This argument is not a guarantee for arbitrary unseen categories.

\subsection{Margin Preservation under Semantic Variation}
\label{sec:margin_open_generalization}

Building on \cref{sec:suppl_positive_margin}, the target-to-distractor
margin for pixel $i$ is
\begin{equation}
    M_i(Z)
    =
    Z_{i,y_i}
    -
    \log
    \sum_{n\neq y_i}
    \exp(Z_{i,n}).
    \label{eq:ov_margin}
\end{equation}
Starting from the standard softmax cross-entropy, we obtain
\begin{equation}
\begin{aligned}
    \ell_i(Z,Y)
    &=
    -\log
    \frac{\exp(Z_{i,y_i})}
         {\sum_{n=1}^{N}\exp(Z_{i,n})}
    \\
    &=
    \log
    \left(
        1+
        \frac{\sum_{n\neq y_i}\exp(Z_{i,n})}
             {\exp(Z_{i,y_i})}
    \right)
    =
    \log
    \left(
        1+\exp(-M_i(Z))
    \right).
    \label{eq:ov_margin_loss}
\end{aligned}
\end{equation}
Since
\begin{equation}
    \frac{\partial \ell_i}{\partial M_i}
    =
    -\frac{1}{1+\exp(M_i)}
    <
    0,
    \label{eq:ov_margin_monotonicity}
\end{equation}
minimizing the perturbed cross-entropy encourages a large
target-to-distractor margin for sampled perturbations.

Let
\begin{equation}
    \widehat{Z}
    =
    A_{\theta}
    \bigl(
        V+\epsilon^{v},
        T+\epsilon^{t}
    \bigr).
    \label{eq:ov_perturbed_logits}
\end{equation}
We use
\begin{equation}
    \mathbb{E}_{\epsilon\sim q_{\phi,\psi}}
    \left[
        M_i(\widehat{Z})
        -
        M_i(Z)
    \right]
    >
    0
    \label{eq:ov_margin_preservation}
\end{equation}
as a diagnostic criterion for positive-incentive behavior. Because the clean
logits $Z$ are not explicitly compared with $\widehat{Z}$ in
\cref{eq:ov_distributional_objective}, this relative-margin condition is
not guaranteed by the practical objective. The training dynamics in the
main paper empirically examine whether target-region alignment improves
relative to the clean features.

The criterion itself is category-agnostic: it evaluates whether a textual
direction supported by the visual evidence remains stronger than competing
directions under plausible visual and textual variations. Consequently, if
an unseen prototype $T_u$ lies near a covered high-probability neighborhood,
for example $T_u\approx T_c+\epsilon_c^t$ with
$\epsilon_c^t\in\mathcal{E}_c(\rho)$, it may reuse the relative
discrimination behavior learned around $T_c$. Deterministic training
supervises margins at isolated seen prototypes, whereas distributional
training samples local neighborhoods around them.

\subsection{Scope of the Interpretation}
\label{sec:ov_scope}

The above argument is an interpretation rather than an unconditional
generalization guarantee. Its benefit depends on three conditions:
the pretrained vision--language model must provide a meaningful
representation of the unseen concept; the unseen prototype should lie
on or near a semantically continuous, high-probability region explored
during training; and the perturbation magnitude must remain controlled
so that semantic neighborhoods do not cross class boundaries. Under
these conditions, Pi-Seg can better preserve and exploit the pretrained
open semantic structure instead of overfitting the downstream model to
fixed seen prototypes. This interpretation is consistent with the
improved unseen-category results reported in
\cref{tab:uavid_classwise,tab:udd5_classwise_seen_unseen}.

\end{document}